\pdfoutput=1
\documentclass[12pt]{article}
\usepackage{graphicx}
\usepackage{subfigure}
\usepackage{amsmath,amssymb} 
\usepackage{color}
\usepackage{bm}
\usepackage{multirow}
\usepackage{url}
\usepackage[letterpaper, margin=1in]{geometry}

\begin{document}

\title{Action4D: Real-time Action Recognition in the Crowd and Clutter} 

\author{Quanzeng You and Hao Jiang \\ Microsoft, USA }

\date{}

\maketitle

\begin{abstract}
Recognizing every person's action in a crowded and cluttered environment is a challenging task.
In this paper, we propose a real-time action recognition method, Action4D, which gives
reliable and accurate results in the real-world settings.
We propose to tackle the action recognition
problem using a holistic 4D ``scan'' of a cluttered scene to include every
detail about the people and environment. 
Recognizing multiple people's actions in the cluttered 4D representation is a new problem.
In this paper, we propose novel methods to solve this problem. 
We propose a new
method to track people in 4D, which can 
reliably detect and follow each person in real time.
We propose a new deep neural network, the Action4D-Net,  
to recognize the action of each tracked person.
The Action4D-Net's novel structure uses both the global feature and the focused attention
to achieve  state-of-the-art result.   
Our real-time method is invariant to camera view angles, resistant to clutter and able to 
handle crowd.   
The experimental results show that the proposed method is fast, reliable and accurate. 
Our method paves the way to action recognition in the real-world applications and is ready to be deployed  
to enable smart
homes, smart factories and smart stores.      
\end{abstract}

\section{Introduction}
Action recognition is a key task in computer vision. Even though human vision 
is good at recognizing subtle actions, computer vision algorithm still cannot achieve the same robustness and accuracy. 
The difficulty is largely caused by the variations of the visual 
inputs. The input video may be crowded and cluttered. 
People may have different clothing, different body shapes and are
highly articulated. They often perform the same action in slightly different ways.
The camera viewing angles can be drastically different so that the same action in the training videos may look quite different
from those in the testing videos.

\begin{figure}[tb]
\centering
\subfigure[]{\includegraphics[width=0.35\linewidth]{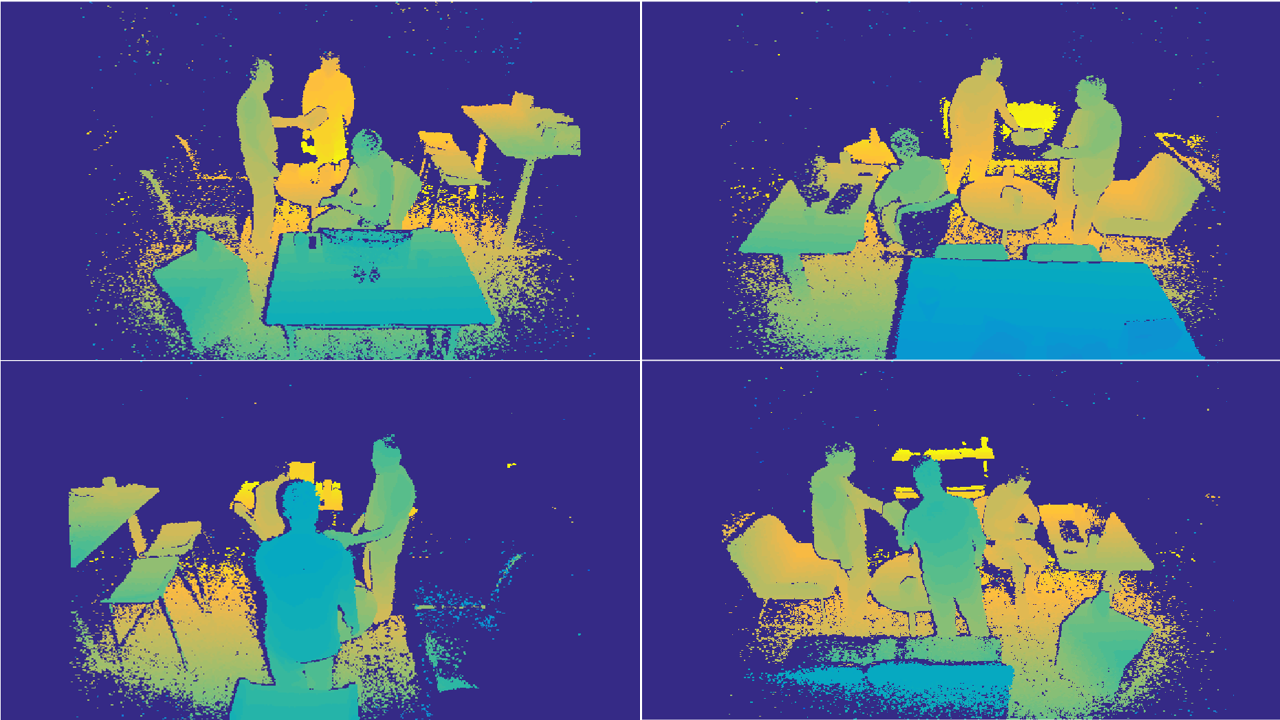}}%
\subfigure[]{\includegraphics[width=0.35\linewidth]{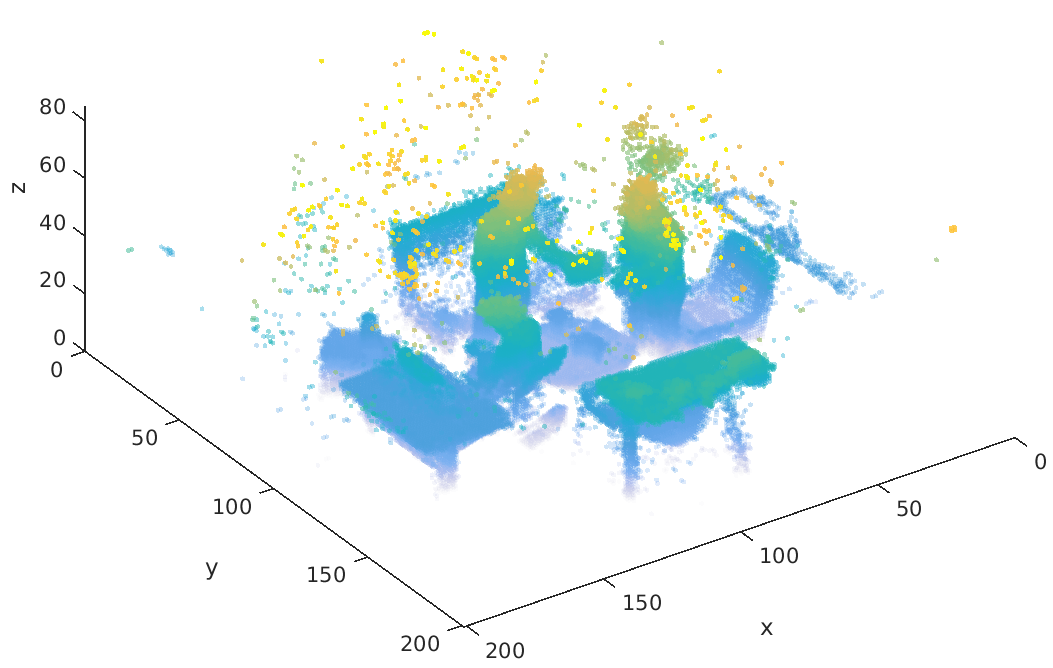}}%
\linebreak
\subfigure[]{\includegraphics[width=0.35\linewidth]{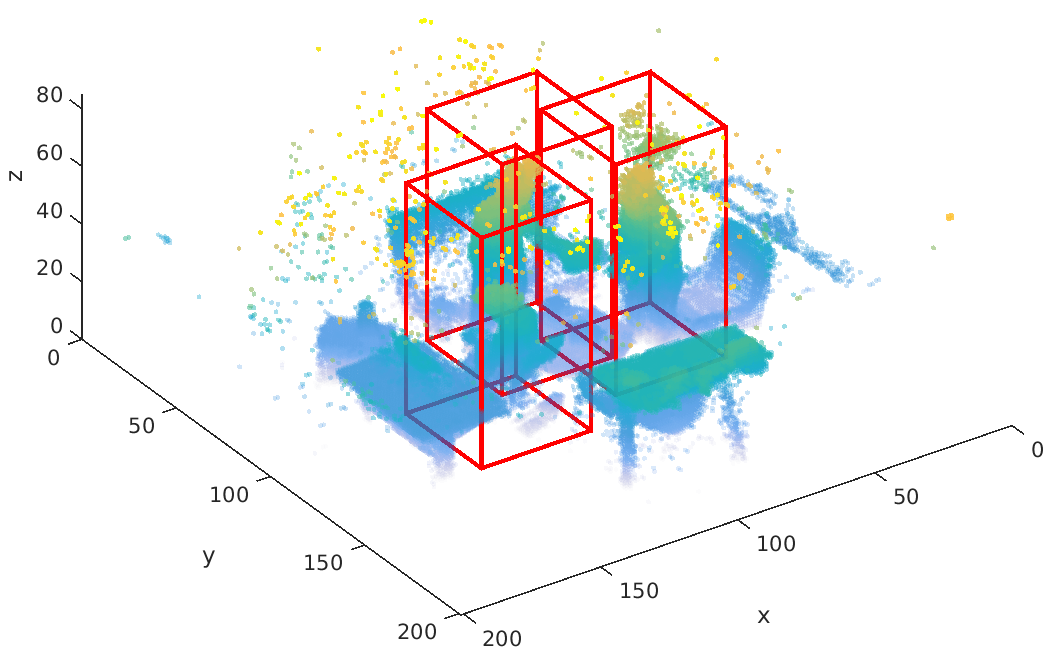}}%
\hspace{20pt}\subfigure[]{\framebox{\includegraphics[width=0.27\linewidth]{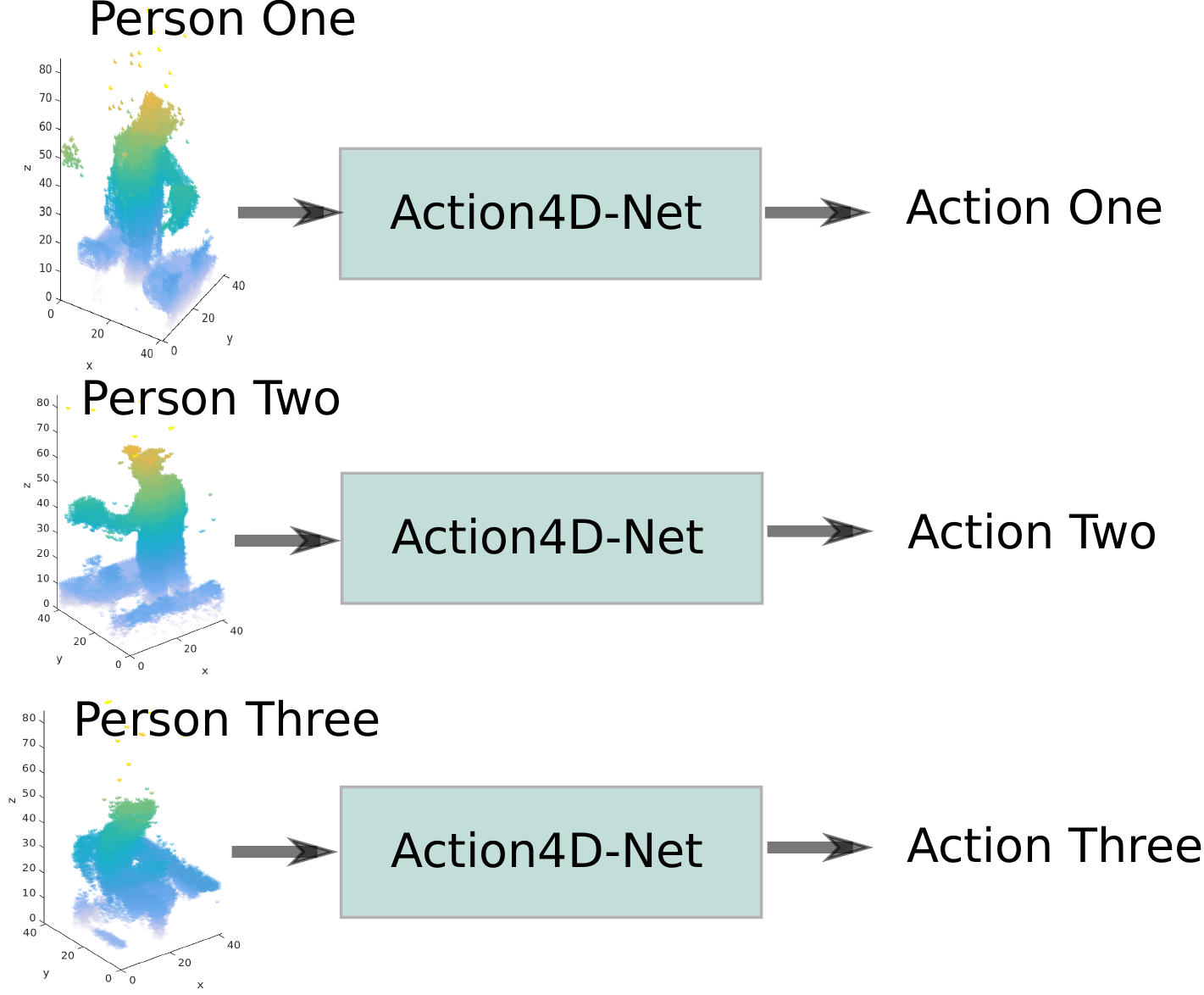}}}%
\vspace{-10pt}
\caption{(a): Depth images from four RGBD cameras. (b): The generated solid model of the scene. 
The 4D solid representation includes people and background context.
(c): Bounding boxes of the tracked people in a solid volume. 
(d): Each cropped solid model is
sent to our Action4D-Net for action recognition.  
Our real-time method is able to handle clutter and crowd. 
}
\label{fig:teaser}

\end{figure}

In this paper, we propose a novel 4D method for robust action recognition.
The input of our method is a 4D solid modeling of the dynamic environment
and our method tracks the action of each person in the cluttered scene.
Recognizing multiple people's actions in cluttered 4D space is a new problem.
To our knowledge, our method gives the first solution to this problem.   
Fig.~\ref{fig:teaser} illustrates our scheme. The proposed method tracks each individual person 
using the 4D solid representation and recognizes their actions in real time.
It is also view invariant, is able to handle crowd and clutter,
and is scalable to applications in a huge space with hundreds of cameras.  


\subsection{Related Works}
In previous studies, most action recognition methods work on single view 2D videos.
Accumulated foreground shape \cite{kidsroom} has been used to recognize the actions in the KidsRoom project. 
In \cite{greg}, shape context is used to model the whole body configuration in action recognition.
Apart from RGB color, motion is also a useful feature for action recognition \cite{efros}.
Other popular handcrafted features for action recognition include spatial-temporal features \cite{kth} and spatial-temporal volumes \cite{irani1}. Based on these features, action detection and recognition can be formulated as a matching problem.
By careful design, we do not even need to directly extract the features;    
the space and time matching can be efficiently solved using low rank analysis \cite{irani2}.

In recent years, deep learning approaches have been widely used in action recognition and detection on color 
images \cite{actiontube,threestream,anothericcv17,3dnetwork}. 
These deep learning methods use multiple streams such as color, motion, body part heat map and 
find actions in the spatial-temporal 3D volume. 
Single view depth images have also been used in action recognition \cite{depth1}.
However, training classifiers for action recognition using 2D images is a challenging task.
It requires training data to include all kinds of variations about camera settings, 
people clothing, object appearances, 
and backgrounds.

Most of the current 3D action recognition methods depend on Kinect 3D skeleton extraction, which can relieve the view dependency
issue in 2D action recognition.
Unfortunately, Kinect skeleton estimation becomes unreliable in cluttered environments. In addition, 3D skeletons alone are insufficient for action recognition. For instance, disambiguating actions such as playing with phone in hands and reading a book is tricky without knowing the objects in people's hands. 
3D people volume from visual hull \cite{visualhull} has also
been extensively used in action recognition \cite{vh1,vh2,vh3}.
Traditional 
visual hull methods usually need special blue/green
or static background and background subtraction to single out people from the background. 
This extremely limits its application in the real-world situation. Figure/ground separation is still a hard problem. State-of-the-art semantic segmentation methods
often miss body parts, mis-classify background objects to foreground people, or
miss details such as the objects people are interacting with.

In contrast, our 4D volume 
representation contains information about not only people but also the objects they are interacting with. 
Without the dependency on people segmentation, 
our method can be robustly applied to crowded and cluttered environments. Traditional methods assume clean segmentation of each individual person in a volume and usually use
handcrafted features such as moments and shape context \cite{shapecontext}. Instead, we propose a novel deep learning approach to handle the cluttered 4D inputs.

Fusion4D \cite{fusion4d}, designed for motion capture, is related to our method. It requires special hardware for high quality stereo to reconstruct detailed mesh representations of the foreground objects. It also uses complex optimization method
for accurate point alignment. In contrast, we build solid representation from noisy depth data, and our task is action recognition.    

To recognize each person's action, we need accurate people tracking.
Multiple camera people tracking  
has been intensively studied. Most of the previous methods use background subtraction to remove the background clutter. 
Unfortunately, background subtraction or figure/ground separation is hard for unconstrained dynamic environment.
Our 4D tracker does not need figure/ground separation and is able to work on the noisy 4D data directly.      
K-shortest path tracker (KSP) \cite{ksp} is related to our approach. However, KSP tracker and our method are quite different.  
KSP uses background subtraction and its occupancy map is on the 2D ground plane, while our approach works directly on the noisy 4D volumes.
Our tracking graph includes prediction nodes and can handle missing detections, while KSP cannot.     

In summary, we propose a novel method to recognize the action of each subject 
in a cluttered environment using real-time 4D solid modeling.
Our work has the following contributions:
\begin{itemize}
 \item  We tackle the new problem of recognizing multiple people's actions in cluttered 4D volume data.
 \item  We propose a new low complexity method to generate dynamic 4D solid volume data that scans all the details in a scene.
 \item  We propose a new long-term people detection and tracking method using the 4D solid volume data.
 \item We propose a new deep neural network, Action4D-Net, for action recognition.
 To our knowledge, our approach is the first attempt to apply deep neural networks on cluttered ``holistic'' 4D solid volume data for action recognition.
 \item There is no existing 4D dataset that includes multiple people and clutter. We collected and labeled a new 4D dataset in 
       our experimentation. We will publish the dataset. 
\item Our proposed method is real-time, 
resistant to crowd and clutter, and it 
can be directly used in complex real-world applications.
\end{itemize}

\section{Method}    

\subsection{Real-time Solid Model Generation}
Given a set of calibrated RGBD images, we compute the occupancy of each voxel in the space. In
a world coordinate system, the voxel center coordinates are denoted as $(\bm x_i, \bm y_i, \bm z_i)$, where $i = 1 .. N$.
We have $M$ cameras and their extrinsic matrices are $[\bm{R}_j|\bm{t}_j]$, where $j = 1..M$ and $\bm{R}_j$
is the rotation matrix and $\bm{t}_j$ is the translation vector and their intrinsic matrices are $\bm{K}_j$.
The depth images
from these cameras are denoted as $\bm{D}_1, ..., \bm{D}_M$. In the following we use $0$ and $1$ to represent false and true respectively.
The occupancy of the voxel at $(\bm  x_i, \bm y_i, \bm z_i)$ from camera $j$
is computed as
\begin{equation}
    \bm{O}_j(i) = [\bm{R}^3_j | \bm{t}^3_j][\bm x_i, \bm y_i, \bm z_i,1]^T \ge \bm{D}_j(\bm{K}_j[\bm{R}_j | \bm{t}_j][\bm x_i, \bm y_i,\bm z_i,1]^T),
\end{equation}
where $\bm{R}^3_j$ and $\bm{t}^3_j$ are the third row of $\bm{R}_j$ and $\bm{t}_j$.
$\bm{O}_j(i)$ is also conditioned on the camera field of view: if the projection
$\bm{K}_j[\bm{R}_j |\bm{t}_j][\bm x_i, \bm y_i, \bm z_i,1]^T$ is outside of the field of view,
we set $\bm{O}_j(i) = 1$.    
The occupancy $\bm{O}(i)$ of the voxel $i$ is thus the intersection of $\bm{O}_j(i)$ from all the $M$ cameras:
\begin{equation}
   \bm{O}(i) = \cap^M_{j=1}\{\bm{O}_j(i)\}.
 \end{equation}
Note that we only carve out the volume we can see from that camera. This is a critical step
that allows us to construct solid modeling even when depth cameras' field of views have no overlap.

The above method is usually sufficient. We adopt the following two techniques to further improve the quality of solid volumes. 1) We use the orthographic top-down view of the point cloud in the volume as a mask to remove the small ``tails" introduced at camera boundary regions.
2) We take steps to deal with small mis-synchronization among cameras. 
Poor synchronization among cameras can lead to the vanishing of thin structures in the 4D volumes. We use the best-effort fashion 
to extract frames from all the Kinect sensors linked together into a local network. 
For fast moving body parts, such as arms,
small mis-synchronization may occur. To remedy this issue, we further inject all the points from depth cameras into 
the solid volume: we set $\bm O(i)$ to one if there is a point in the voxel $i$. These voxels are 
on the scene surface and the other ones are internal voxels.         
Fig.~\ref{fig:teaser}~(b) shows an example of the solid volume data.
The holistic property of the 4D volume is the key to reliable action recognition. 
Beyond the volume occupancy, each voxel can have other attributes such as the RGB color.  In this work, for action recognition we ignore the color and only consider the occupancy of each voxel.
 
Directly computing the 4D solid volumes using CPU is slow due to the large number of voxels, \textit{e.g.} a volume of dimension $201\times201\times85$ is constructed in our experimental environment.
Fortunately, we can pre-compute all $[\bm R^3_j | \bm t^3_j][\bm x_i, \bm y_i, \bm z_i,1]^T$ and $\bm K_j[\bm R_j | \bm t_j][\bm x_i, \bm y_i, \bm z_i,1]^T$.
Then, we can easily parallelize  all the comparison operations in GPU. Similarly, the point cloud filling and top-down
carving can also be done in parallel. Currently,  our implementation enables real-time performance with only 1 to 2 percent of GPU usage.
      
\subsection{People Detection and Tracking}
Given a 4D volume that is a complete scan of a cluttered scene, direct action recognition would be tricky. Detecting each subject
in the scene is necessary, so our attention can be focused on each individual. 
For action recognition, we also need to observe every subject in a duration.
We thus need to track each subject in the scene. Tracking also helps remove false people detections and smooth out missing
ones. 

\subsubsection{Candidate detection}
Even though a sweeping volume solution is feasible to detect object candidates, it has high complexity. 
We use a light-weight solution here: 
we work on the top-town envelop image. Let $f(m,n,k)$ be the volume data and $m,n$ are $x,y$ coordinates
and $k$ is the $z$ coordinate. we assume $z=0$ is the ground plane. The top-down envelop is  
$g(m,n) = \mbox{max}_k(\phi(f(m,n,k)))$, where $\phi(f(m,n,k)) = k$ if 
$f(m,n,k) > 0$ and otherwise $\phi(f(m,n,k)) = 0$. Based on the observation that each potential object corresponds to at 
least one local maximum on $g$, we use a simple Gaussian filter to extract the candidates. The local maxima are found
on the Gaussian filtered top-down envelop using non-maximum suppression. 
Each candidate volume is a cuboid around a local maximum with a given width and height. Currently, we set the height to be 
the height of the 4D volume.

\begin{figure}[tb]
	\centering
	\subfigure[]{\framebox{\includegraphics[width=0.65\linewidth]{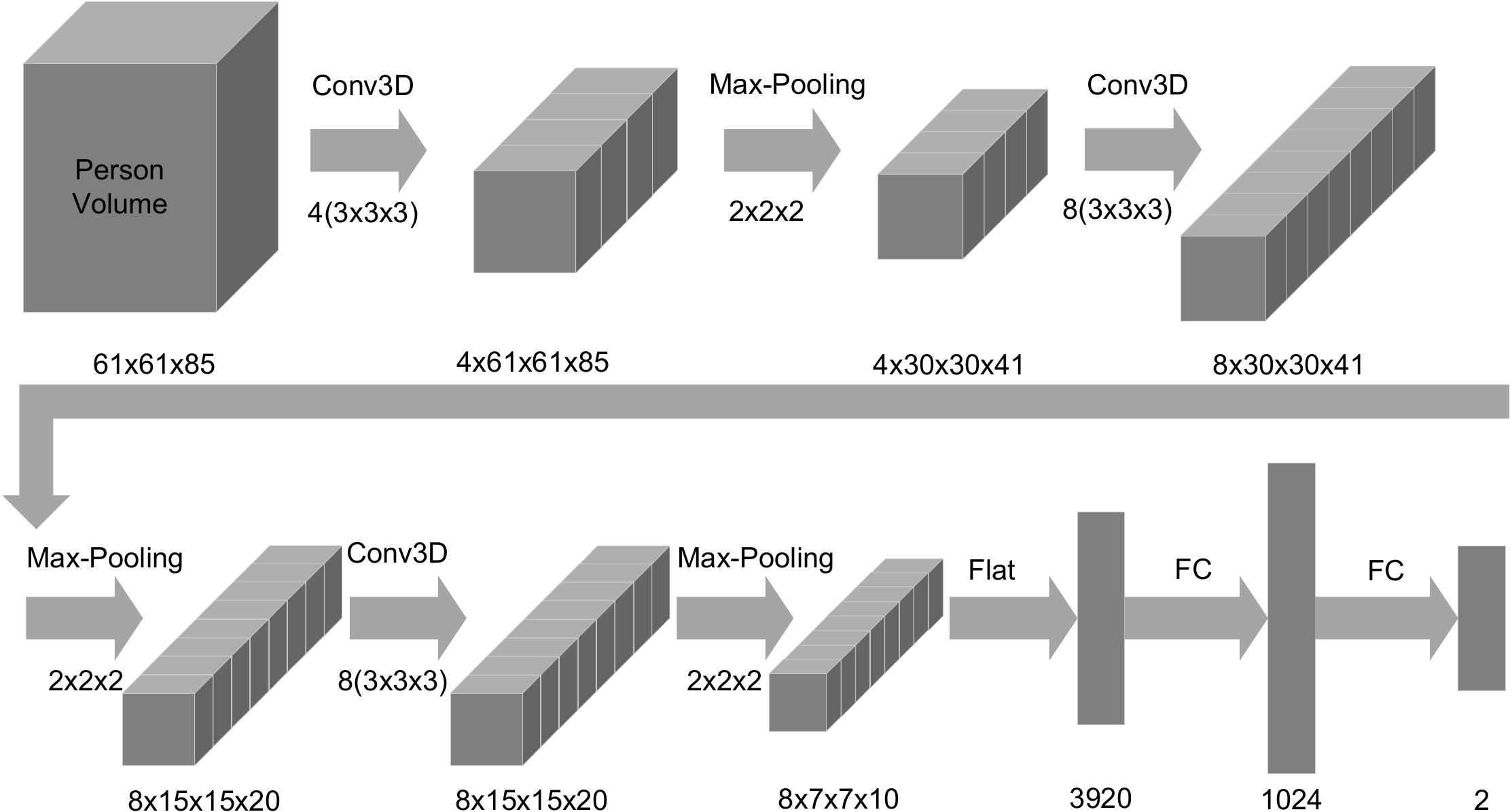}}}%
	\hspace{20pt}%
	\subfigure[]{\framebox{\includegraphics[width=0.218\linewidth]{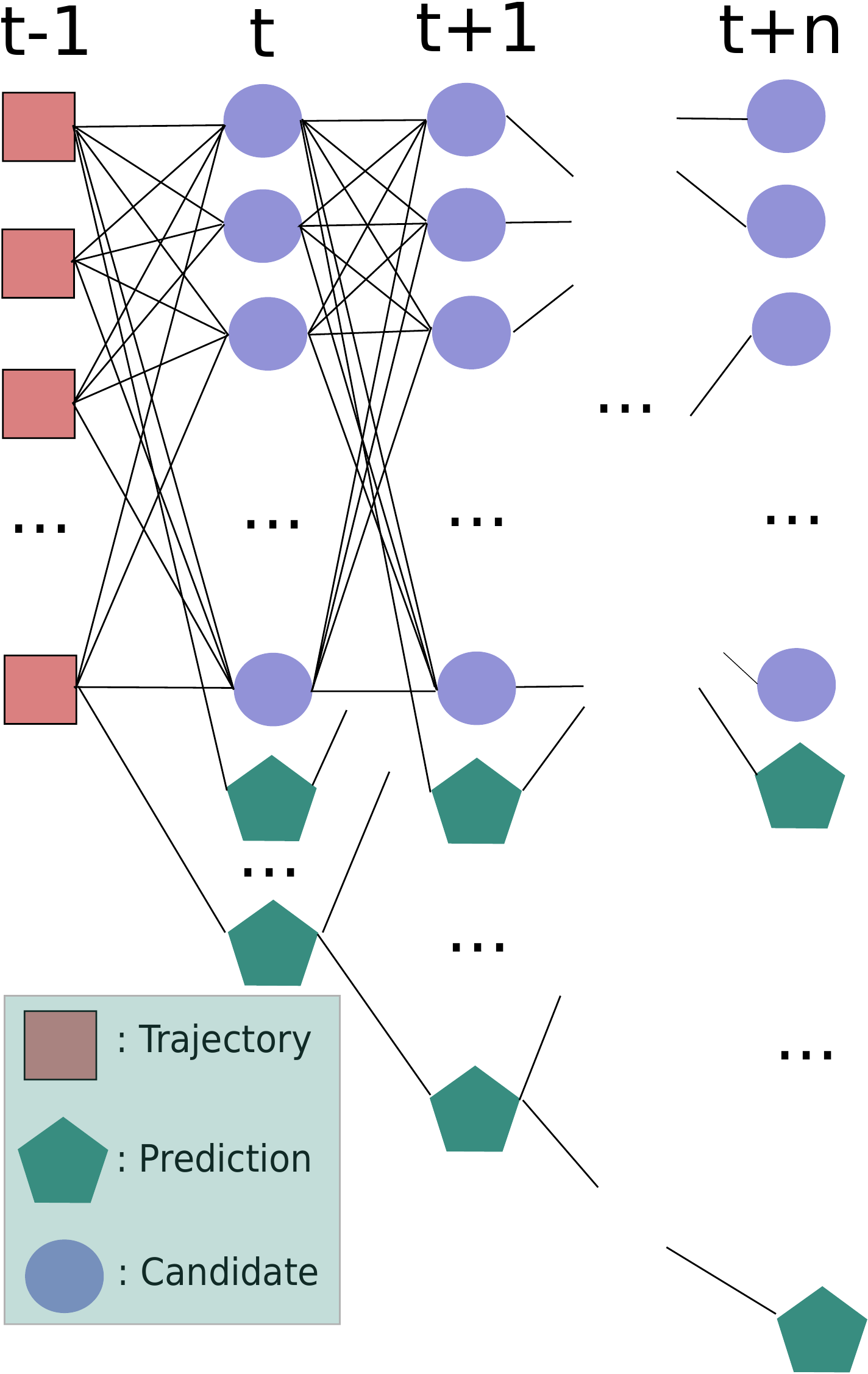}}}%
	\caption{(a): People classification CNN. (b): We find disjoint paths on the tracking graph.}
	\label{fig:peoplenet}
\end{figure}

\subsubsection{3D people classifier}
We train a 3D CNN to classify each candidate volume to be people or non-people.
Even though color can also be used, in this paper the input to the CNN is only the occupancy volume.
The people classifier CNN structure is
shown in Fig.~\ref{fig:peoplenet}~(a). 
The network contains a sequence of 3D convolution layers, ReLUs and pooling layers to extract features
from the 3D volume (ReLUs are not shown). 
The features are then fed into a fully connected network for people classification.   
The 3D people classifier gives the probability of each candidate 3D bounding box containing a person.
Even with just a few thousand frames of training data, the people detector can achieve high accuracy to support the following people tracker.

\begin{figure}[tb]
	\centering
	\includegraphics[width=0.125\linewidth]{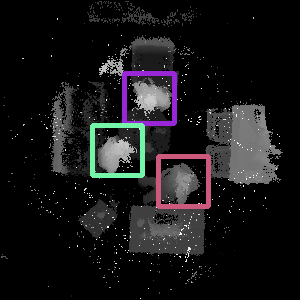}%
	\includegraphics[width=0.125\linewidth]{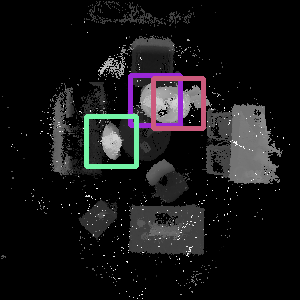}%
	\includegraphics[width=0.125\linewidth]{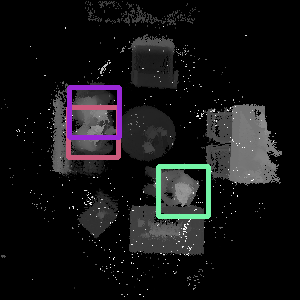}%
	\includegraphics[width=0.125\linewidth]{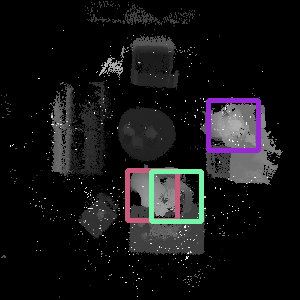}%
	\includegraphics[width=0.125\linewidth]{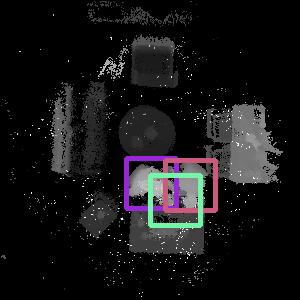}%
	\includegraphics[width=0.125\linewidth]{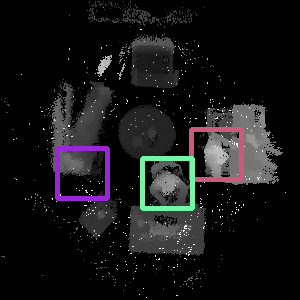}%
	\includegraphics[width=0.125\linewidth]{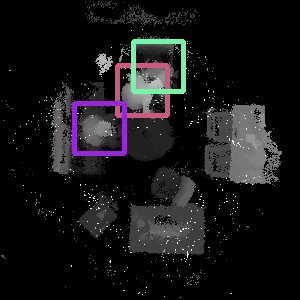}%
	\includegraphics[width=0.125\linewidth]{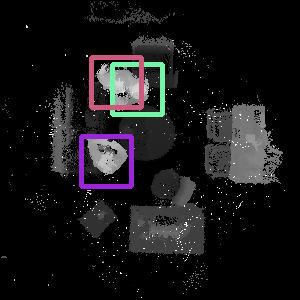}%
	\linebreak
	\includegraphics[width=0.125\linewidth]{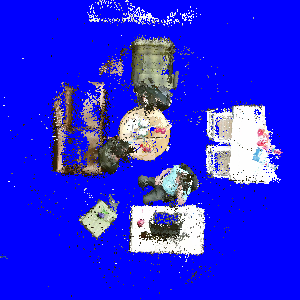}%
	\includegraphics[width=0.125\linewidth]{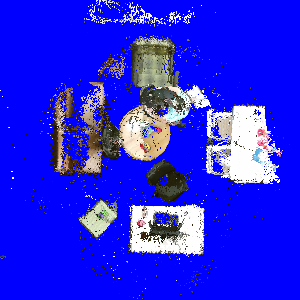}%
	\includegraphics[width=0.125\linewidth]{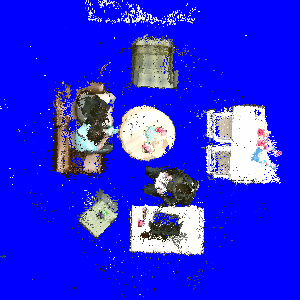}%
	\includegraphics[width=0.125\linewidth]{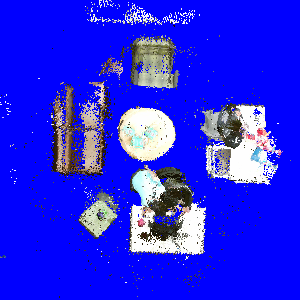}%
	\includegraphics[width=0.125\linewidth]{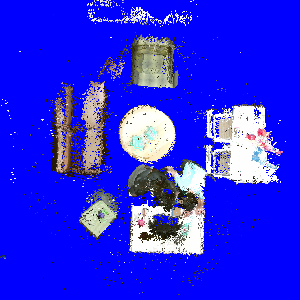}%
	\includegraphics[width=0.125\linewidth]{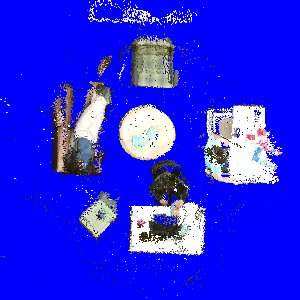}%
	\includegraphics[width=0.125\linewidth]{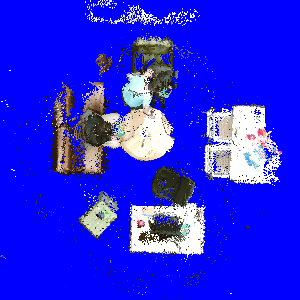}%
	\includegraphics[width=0.125\linewidth]{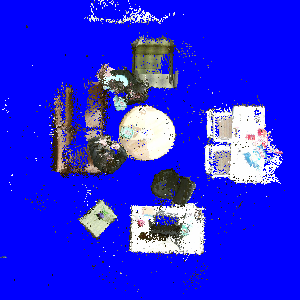}%
	\caption{Sample tracking results of the proposed method.
		Row one: Tracking results overlapped on the top-down envelop of the 4D volume. Each person's bounding box has a different color.
		Row two: The corresponding color images synthesized from the RGB images and the top-down envelop.
		The proposed real-time tracking method is able to distinguish and follow the targets robustly through the  
		video, even when the subjects stay very close to each other.
                (See the video at \protect\url{http://www.hao-jiang.net/projects/action4d/video.html}).
	}
	\label{fig:tracking}
\end{figure}

\subsubsection{Optimizing people tracking}
With the extracted candidates, people tracking can be formulated as a path following problem.
We try to link the detected trajectories to the detections in the current frame $t$ and 
the next $n$ frames. Here $n$ is a small number, \textit{e.g.}, three. The idea is that by introducing a little 
bit delay of $n$ frames. We can achieve much more reliable people tracking.

The tracking graph is shown in Fig.~\ref{fig:peoplenet}~(b). There are three kinds of nodes in the graph: the rectangle nodes represent
the trajectories already formed, the oval nodes represent candidates, and the pentagon nodes are the prediction nodes. The number of prediction nodes 
equals that number of candidate nodes plus the prediction nodes at the previous time instant. 
The edges in the graph indicate the possible matches between nodes.
The edge weights are determined by the difference of the probabilities from the 3D people classifier, the Euclidean distance,
the occupancy volume difference, and the color histogram difference between neighboring nodes. The trajectory node also has a weight inversely 
proportional to the trajectory length.  
To track objects in the scene, we find the extension of each trajectory from time $t-1$ to
$t+n$, so that these paths pass each trajectory node and all the paths are node disjoint.

This optimization problem can be reduced to a min-cost flow problem and
it can be solved efficiently using a polynomial algorithm \cite{mincostflow}.
Each trajectory is only extended to the
neighboring nodes within a radius $d_L$, which is determined by the max-speed
of a person and the frame rate of the tracking algorithm.
The gating constraint also speeds up the optimal path search.

After the optimization, we extend each existing trajectory by one-unit length. We remove
trajectories with low people score, where the people score is defined as the weighted sum 
of the current people probability and the previous people score. And, we include new trajectory for each 
candidate node at time $t$ that is not on any path. The new set of trajectories are used to form
a new graph for the next time instant. We repeat the procedure each time when a new video frame is received.       

Our people detection and tracking algorithm is robust against clutter
and crowd. Fig.~\ref{fig:tracking} shows sample results from our 4D tracking over a few thousand frames. 
The tracker is able to handle cases such as hugging without tracking lose.
\subsection{Action Recognition}

\begin{figure}[tb]
\centering
\subfigure[]{\includegraphics[width=0.125\linewidth]{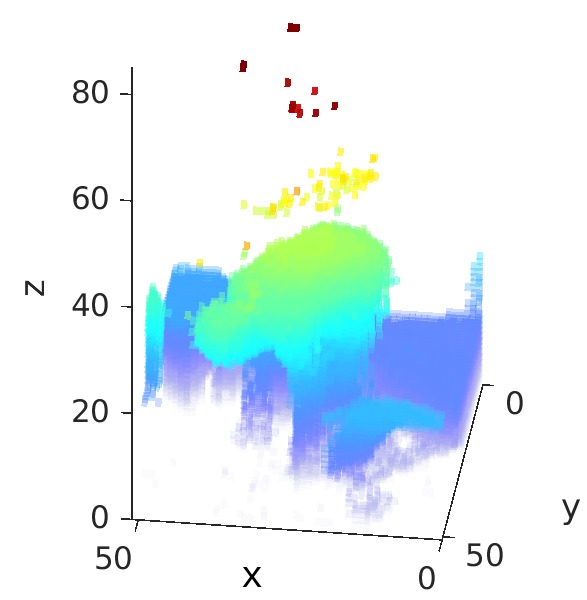}}%
\subfigure[]{\includegraphics[width=0.125\linewidth]{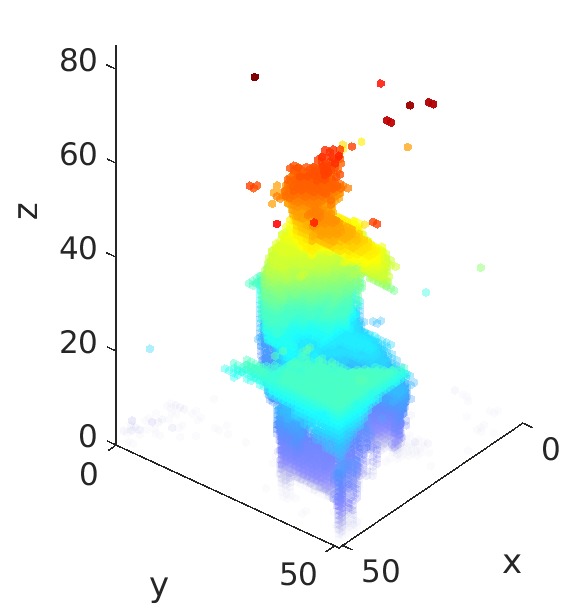}}%
\subfigure[]{\includegraphics[width=0.125\linewidth]{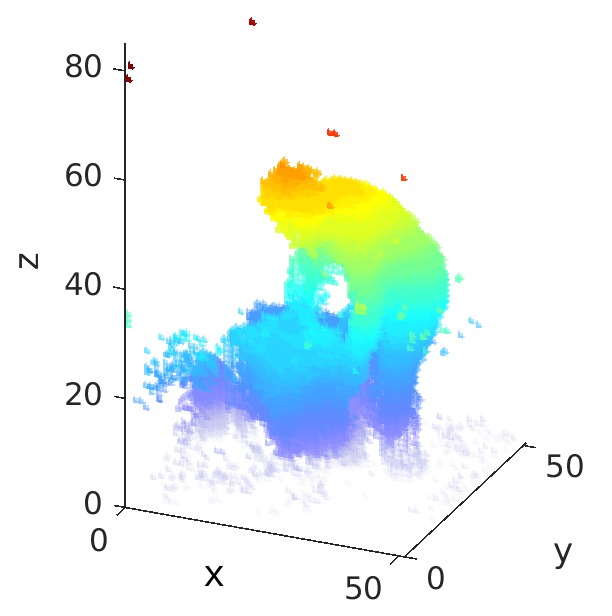}}%
\subfigure[]{\includegraphics[width=0.125\linewidth]{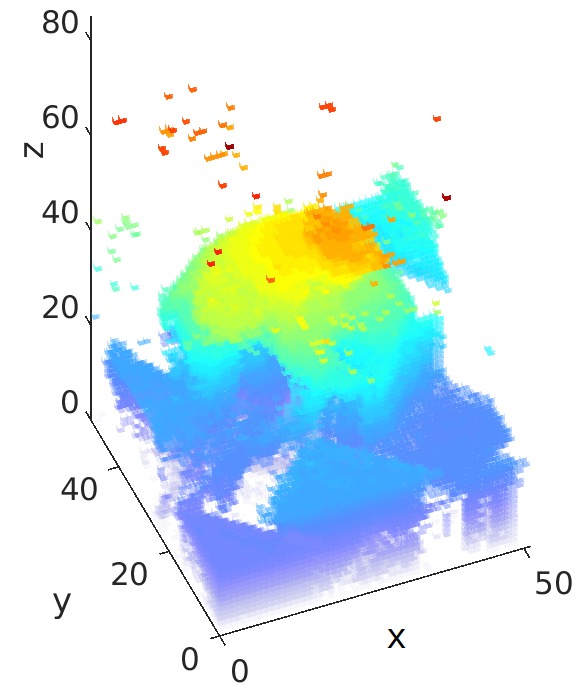}}%
\subfigure[]{\includegraphics[width=0.125\linewidth]{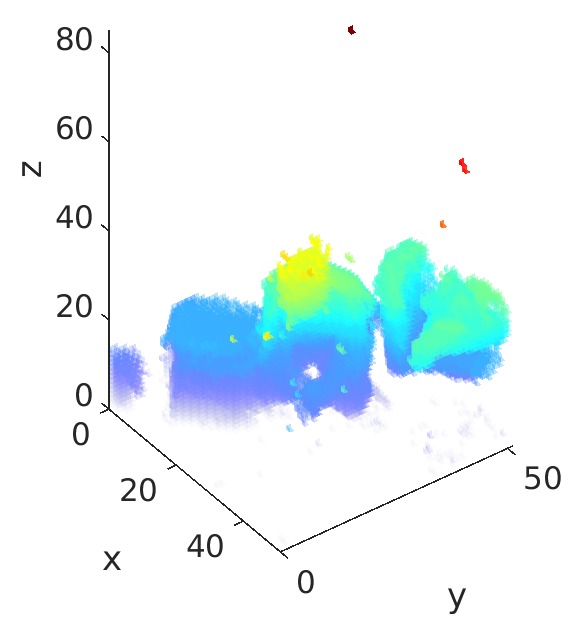}}%
\subfigure[]{\includegraphics[width=0.125\linewidth]{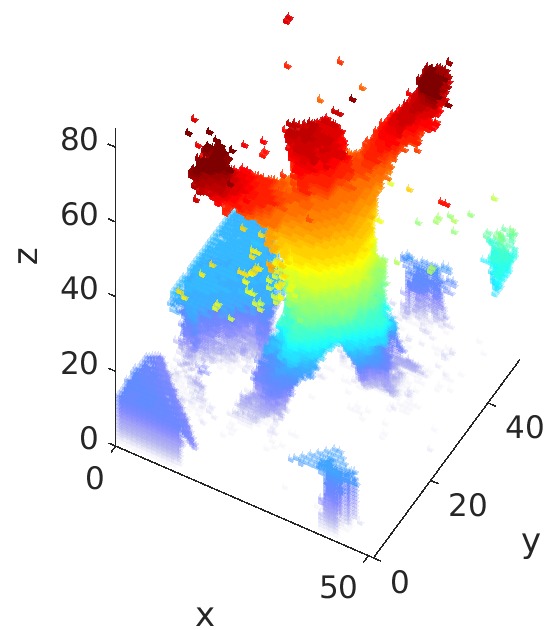}}%
\subfigure[]{\includegraphics[width=0.125\linewidth]{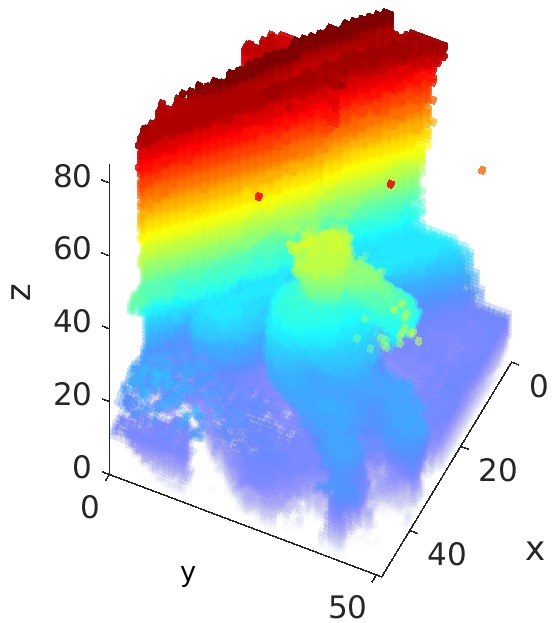}}%
\subfigure[]{\includegraphics[width=0.125\linewidth]{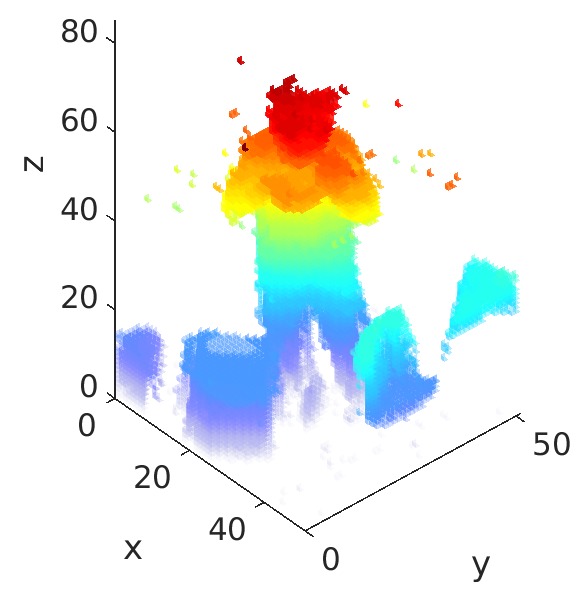}}%
\linebreak
\subfigure[]{\includegraphics[width=0.125\linewidth]{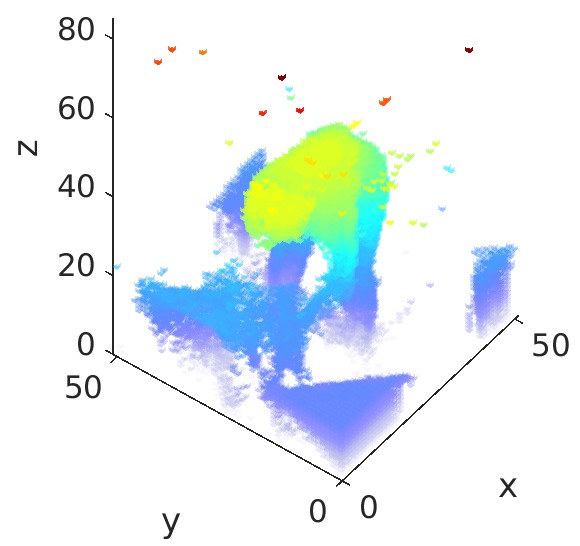}}%
\subfigure[]{\includegraphics[width=0.125\linewidth]{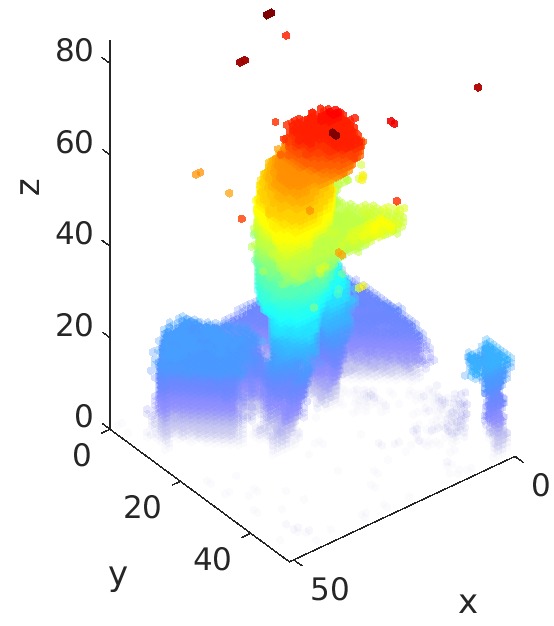}}%
\subfigure[]{\includegraphics[width=0.125\linewidth]{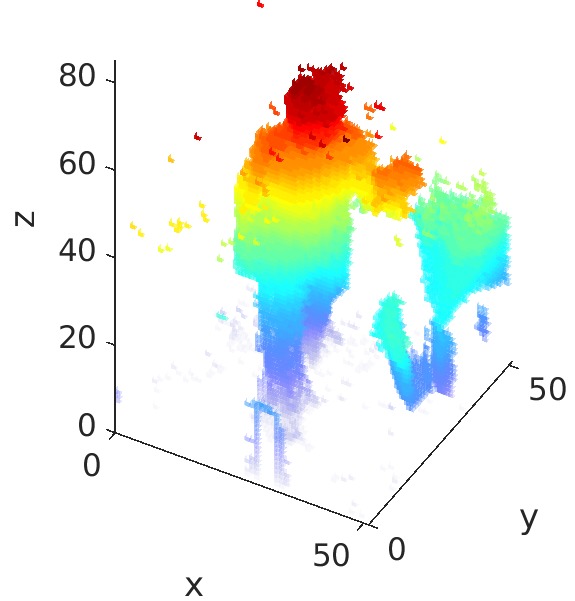}}%
\subfigure[]{\includegraphics[width=0.125\linewidth]{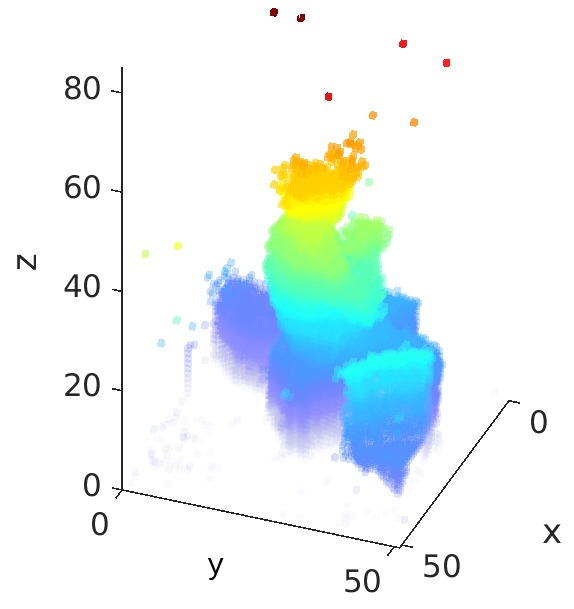}}%
\subfigure[]{\includegraphics[width=0.125\linewidth]{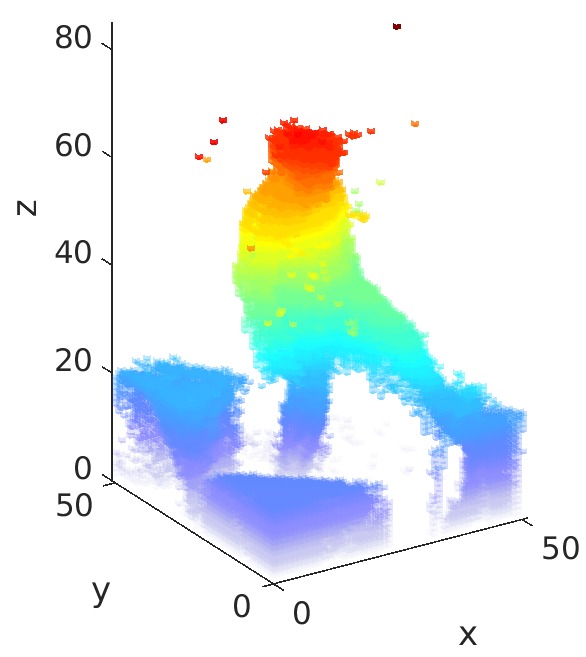}}%
\subfigure[]{\includegraphics[width=0.125\linewidth]{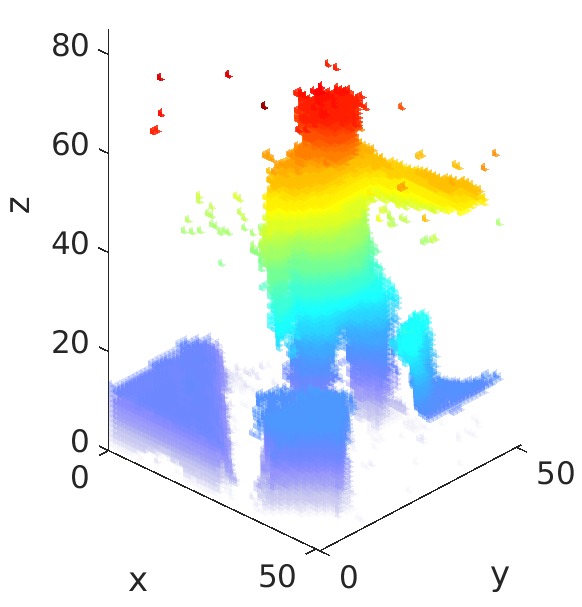}}%
\subfigure[]{\includegraphics[width=0.125\linewidth]{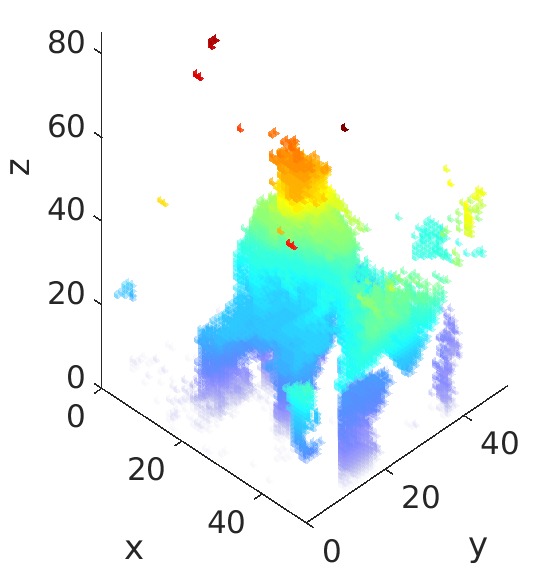}}%
\subfigure[]{\includegraphics[width=0.125\linewidth]{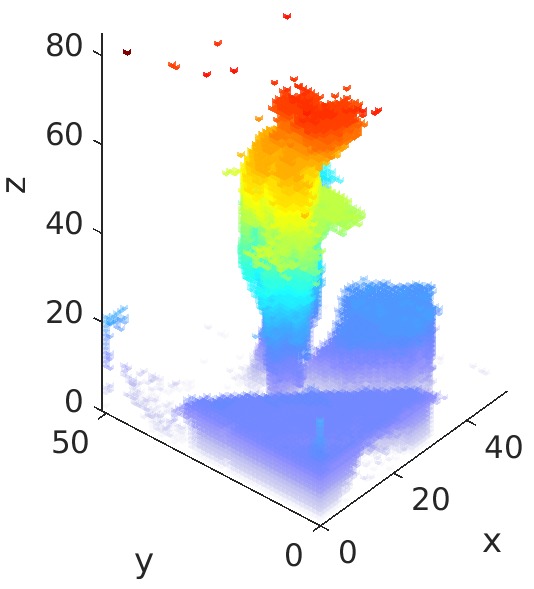}}%
\vspace{-5pt}
\caption{ We recognize actions using solid models.
The action can be easily identified even from a static snapshot.
The actions are (a) bending, (b) drinking, (c) lifting,
(d)pushing/pulling, (e) squadding, (f) yawning, (g) calling, (h) eating, 
(i) opening drawer, (j) reading, (k) waving, (l) clapping, (m) kicking, (n) pointing,
(o) sitting, and (p) browsing cell phone.
These real-time generated
volumes are fed into our Action4D-Net for action recognition.
}
\label{fig:crop}
\end{figure}

Fig.~\ref{fig:crop} shows the cropped solid representation based on the tracker output at different time instants.
As shown in Fig.~\ref{fig:crop},
a person is at the center of each volume and there is often a lot of clutter. 
The solid representation 
clearly shows the action of a person, even without the color.
Also, the objects, which human subjects are interacting with, and the surrounding 
background
are included in the volume. 
This is in fact desirable for action recognition because we can take advantage of all the context information.

In action recognition, we observe the 4D volume data to infer the action at each time instant. 
There are many clues that can be used to infer the action of a person, \textit{e.g.}, the body poses,
the movement of body parts, and the objects the subject is handling. The background context is also
a strong clue. For instance, if we see a chair underneath a person, we can infer that the person is sitting.
By analyzing how a person volume evolves, we are able to recognize the actions.
Potentially, the position or the speed of each person can also be used in action recognition. However, in this paper we depend on the volume data alone for action recognition.

We construct deep convolutional neural network, Action4D-Net, for accurate action recognition.
The inputs are a sequence of person volumes automatically extracted from our people detection and tracking method.
 In the network, the input goes through a sequence of 3D convolution layers
combined with 3D pooling layers to produce action features. 
Meanwhile, we also propose to use an auxiliary attention net, which will be discussed in more details in the following sub-sections.  
These features at each time instant are fed into a Recurrent Neural Network (RNN) to aggregate the temporal information for final action classification.

In the following, we present the network structures in more details.

\begin{figure}[!ht]
        \centering
        \includegraphics[width=0.85\linewidth]{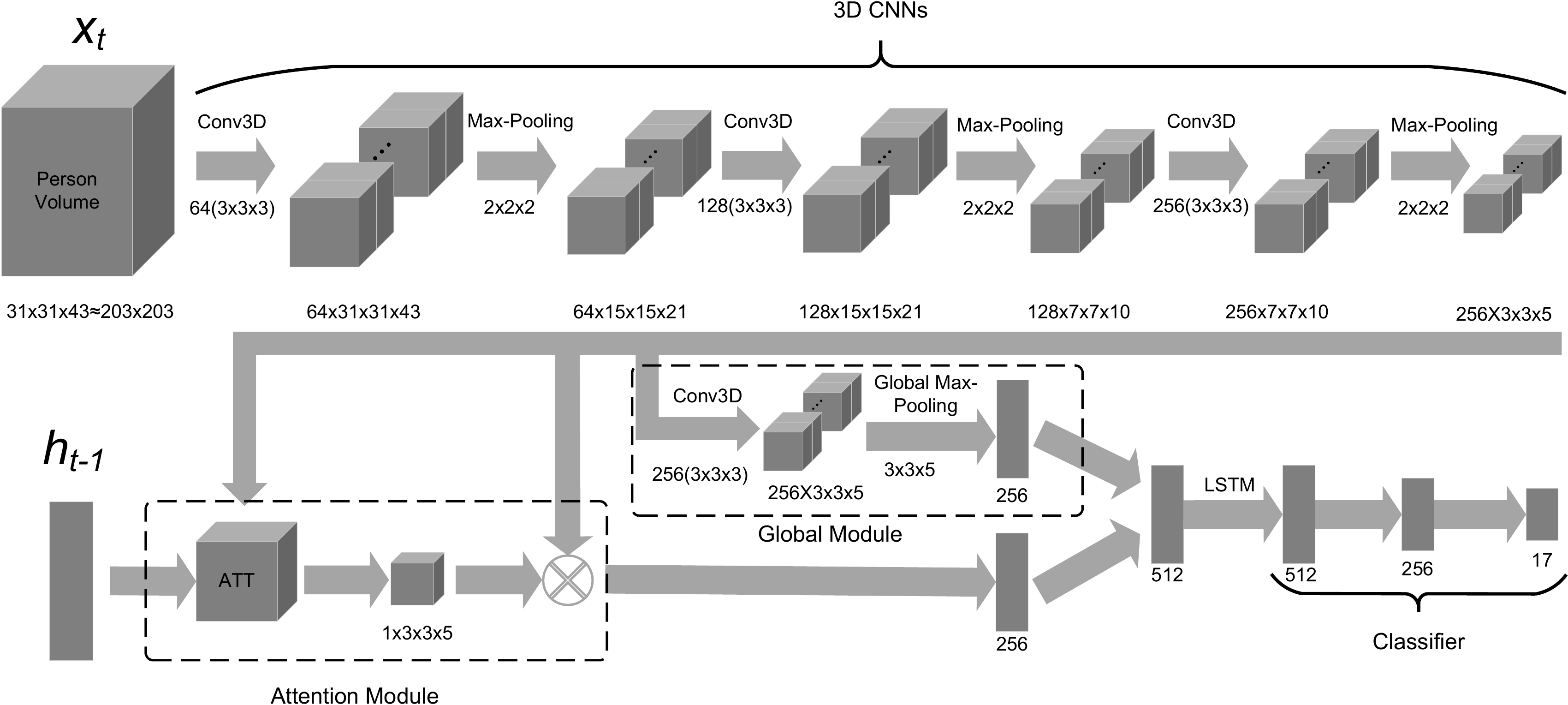}
        \vspace{-5pt}
        \caption{Our proposed attention Action4D-Net for action recognition.}
        \label{fig:model}
\end{figure}
\subsubsection{Attention Action4D-Net}
\figurename~{\ref{fig:model}} shows the proposed neural network architecture for action recognition using the volumetric data.  It starts with several 3D convolution layers followed by 3D max-pooling layers. Then, attention model is employed to generate the local features and global max-pooling~\cite{GlobalPooling} is used for global features. Both the local and the global features are concatenated as the inputs to the Recurrent Neural Network (we use LSTM in this work) to produce the features for action classification.

Intuitively, human beings focus on different regions when recognizing different actions. For instance, when recognizing the action of book reading, 
we mostly focus on the subject's hands and the book in his/her hands. While for drinking water, we change our focus to his/her mouth area. 
To achieve this goal, we use the recently developed attention model, which has been broadly applied to natural language processing and computer vision tasks~\cite{bahdanau2014neural,NIPS2014_5542}. In particular, attention model is able to automatically discover the relevance between different inputs at a given context. In our work, we employ attention model to automatically learn the most relevant local sub-volumes for a given action.

Let $\bm{V}\in \mathbb{R}^{F\times L \times W\times H }$ be the output from the last 3D convolution layer, where $F$ is the number of filters, $L$, $W$ and $H$ are the size of the 3D output. In particular, each location in the  3D output can be represented as $\bm{v}_{ijk}\in \mathbb{R}^{F}$ for  $ 1\le i \le L$, $1 \le j \le W$ and $1\le k \le H$.  The attention weights for all $\bm{v}_{ijk}$ are computed as
\begin{align}
\label{eqn:alpha}
\bm \beta_{ijk} & = \bm{h}_{t-1}^T \bm{U} \bm{v}_{ijk}  \\
\bm{\alpha} & = \mathtt{softmax}(\bm{\beta}),
\end{align}
where $\bm\alpha \in \mathbb R^{L \times W\times H}$ is the attention weights, $\bm U \in \mathbb{R}^{D\times F}$ is the weight matrix to be learned and $\bm h_{t-1}\in \mathbb{R}^{D}$ is the previous hidden state of size $D$ from the Recurrent Neural Network. In such a way, the network is expected to automatically discover the relevance of different sub-volumes for different actions.

Next, the local feature $\bm v$ are computed as the weighted sum of all the sub-volume features $\bm{v}_{ijk}$
\begin{equation}
\label{eqn:attin}
\bm{v} = \sum_{i,j,k} \bm{\alpha}_{ijk} \bm{v}_{ijk} .
\end{equation}

Since there is no guarantee that the attention model can always correctly discover the most 
relevant local body parts for each action, we employ the global max-pooling to extract the 
global feature as extra information for action recognition. For instance, the volumes of 
sitting and kicking look quite different (see Fig.~\ref{fig:crop}), which can be captured by 
the global features of the volumes. We employ a 3D convolution layer followed by a global pooling layer to obtain the global feature $\bm g$ (see Fig.~\ref{fig:model}). Next, both the global  feature $\bm g$ and the local attention feature $\bm v$ are supplied to the LSTM cell to capture the temporal dependencies. The action classification model, which is a multi-layer perceptron (MLP), takes the hidden state from the LSTM cell as input to recognize different actions. 

\section{Experimental Results}
In this section, we evaluate the proposed 4D approach for action recognition
and compare our approach against different competing methods. 

\subsection{Should We Keep Everything in Our Solid Representation?}
In our experimentation, we extract the 4D volume representation for each person 
based on our 4D people tracker. Given each person's location, we extract a volume of $31\times 31 \times 43$, 
whose voxel has the size
$5 cm \times 5cm \times 5cm$. This volume is set large enough to cover a person with different poses. 
Apart from the target subject, background clutter and other subjects in the scene are
also included in the cropped volume as shown in 
Fig.~\ref{fig:crop}. 
	Potential approaches, such as semantic segmentation and 3D skeleton estimation,     
	can be used to separate a person from the clutter. 
	However, should we depend on these methods in action recognition?
	In the following, we show that both of these methods may fail and in fact we want to keep the ``clutter'' because
	it forms the context of actions and is often crucial to correct action recognition.  
	
\begin{figure}[tb]
		\centering
		\includegraphics[width=0.2\linewidth]{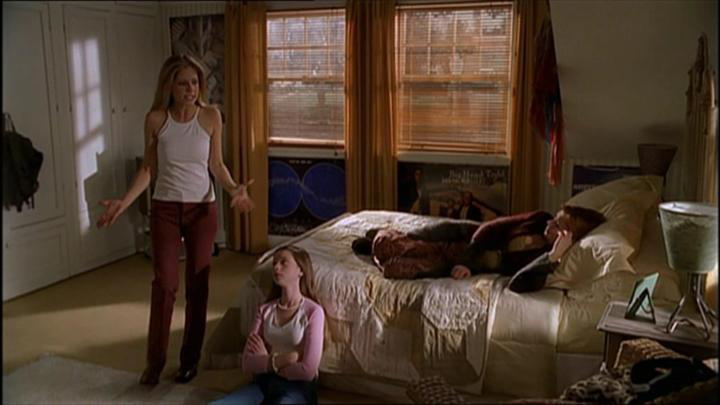}%
		\includegraphics[width=0.2\linewidth]{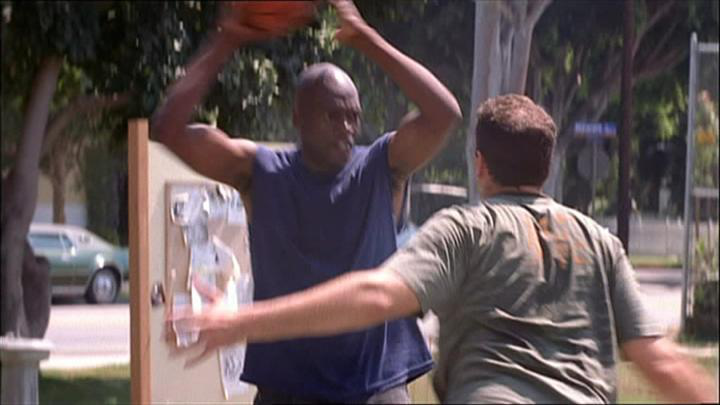}%
		\includegraphics[width=0.2\linewidth]{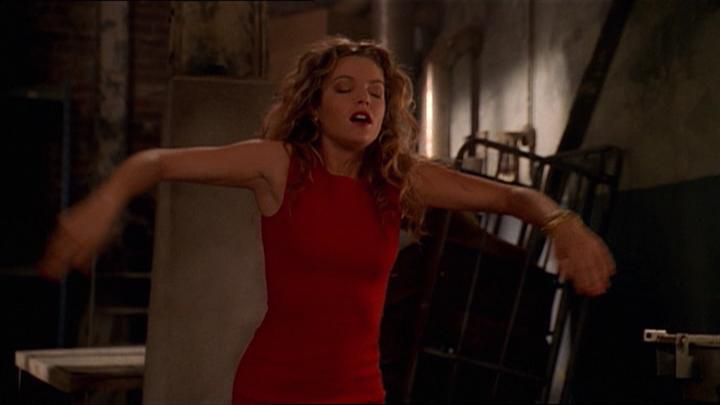}%
		\includegraphics[width=0.2\linewidth]{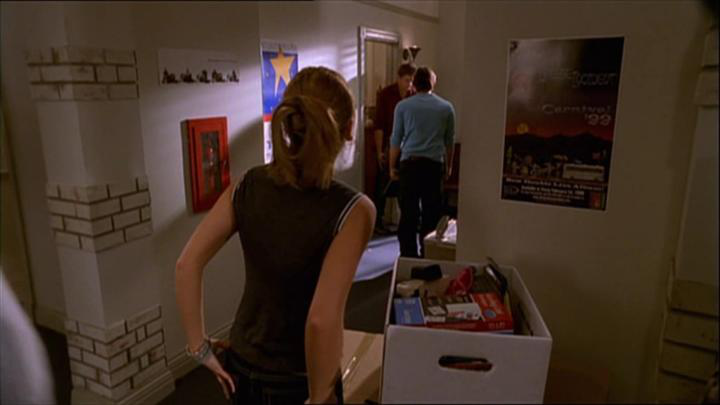}%
		\includegraphics[width=0.2\linewidth]{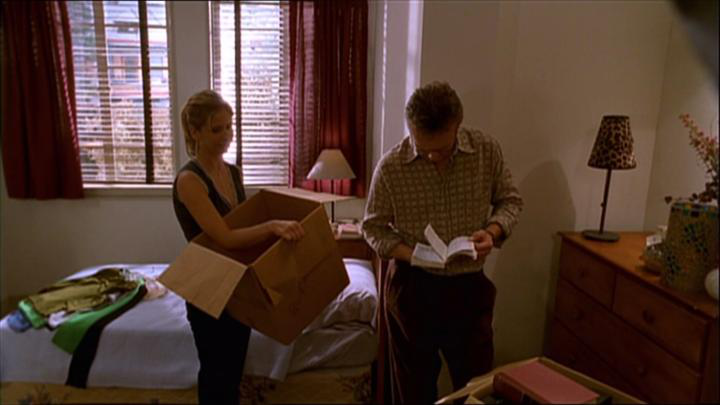}%
		\linebreak
		\includegraphics[width=0.2\linewidth]{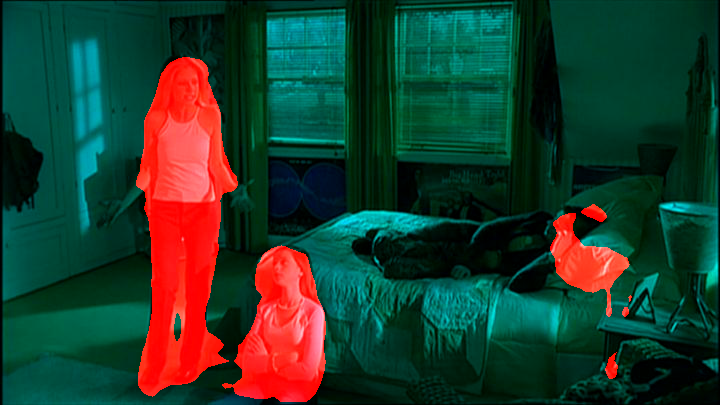}%
		\includegraphics[width=0.2\linewidth]{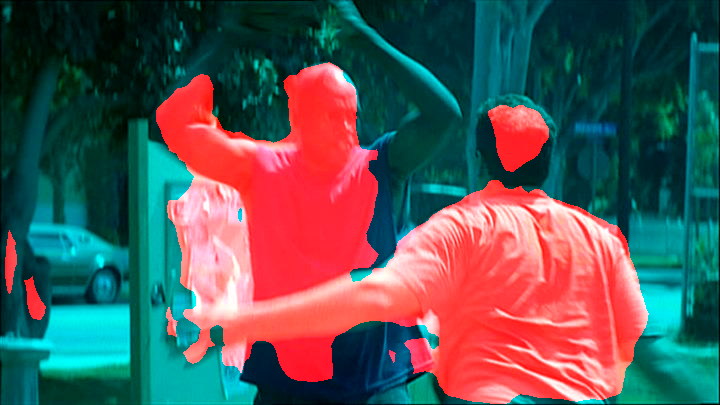}%
		\includegraphics[width=0.2\linewidth]{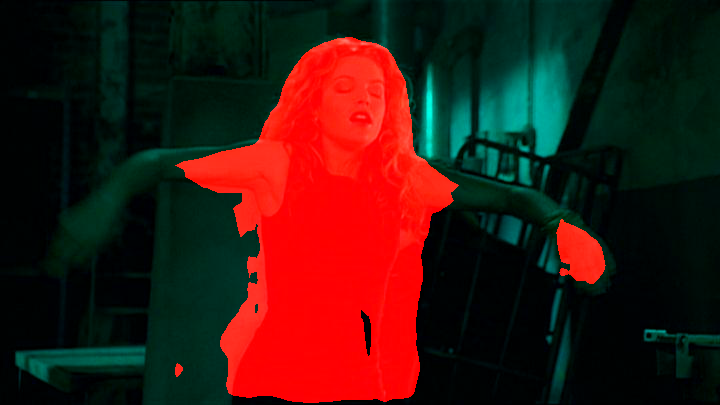}%
		\includegraphics[width=0.2\linewidth]{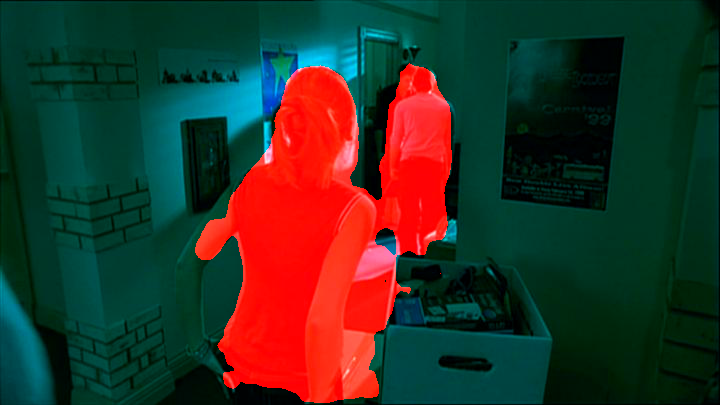}%
		\includegraphics[width=0.2\linewidth]{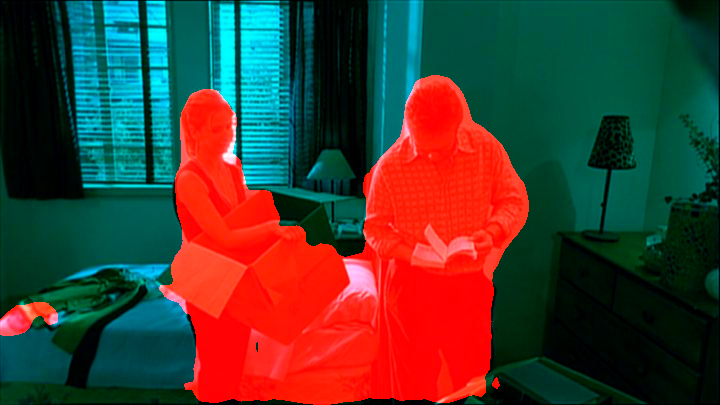}%
		\vspace{-5pt}
		\caption{Semantic segmentation using FCN-8s \cite{fcn} on several images from the Buffy dataset \cite{buffy}. The red mask gives the people segmentation.}
		\label{fig:fcn}
		
	\end{figure}

	Semantic segmentation is not always reliable. In Fig.~\ref{fig:fcn}, we show the semantic segmentation results using the FCN-8s \cite{fcn}, 
	one of state-of-the-art semantic segmentation methods, to separate people from background in the Buffy datatset \cite{buffy}. 
	As shown in Fig.~\ref{fig:fcn}, 
	the people segmentation may lose a person's body parts or include other background objects.
	If we use semantic segmentation as a pre-processing step for action recognition, the segmentation errors will 
	fail the succeeding action recognition algorithm. 
	
	\begin{figure}[tb]
		\centering
		\includegraphics[width=0.125\linewidth]{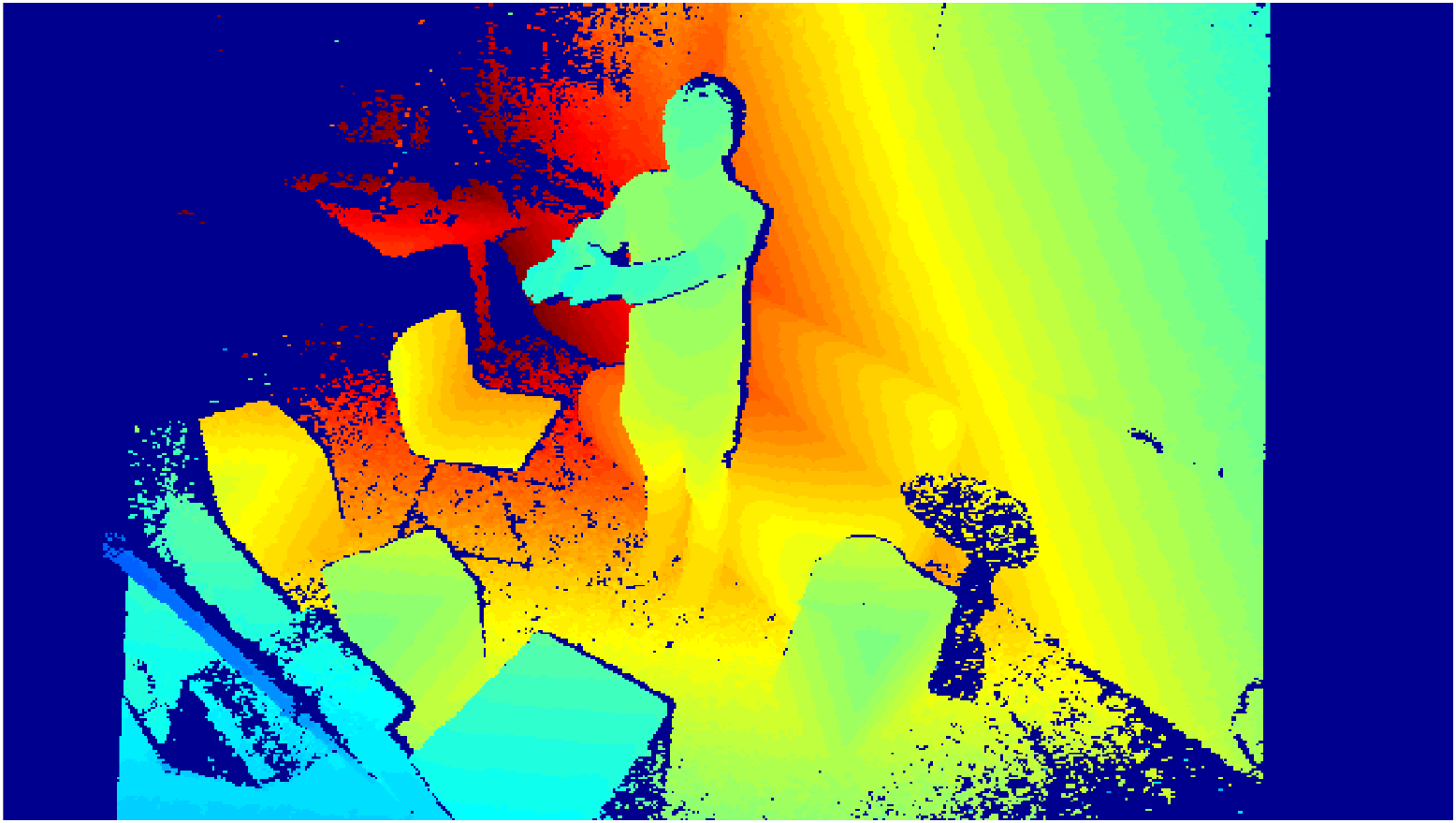}%
		\includegraphics[width=0.125\linewidth]{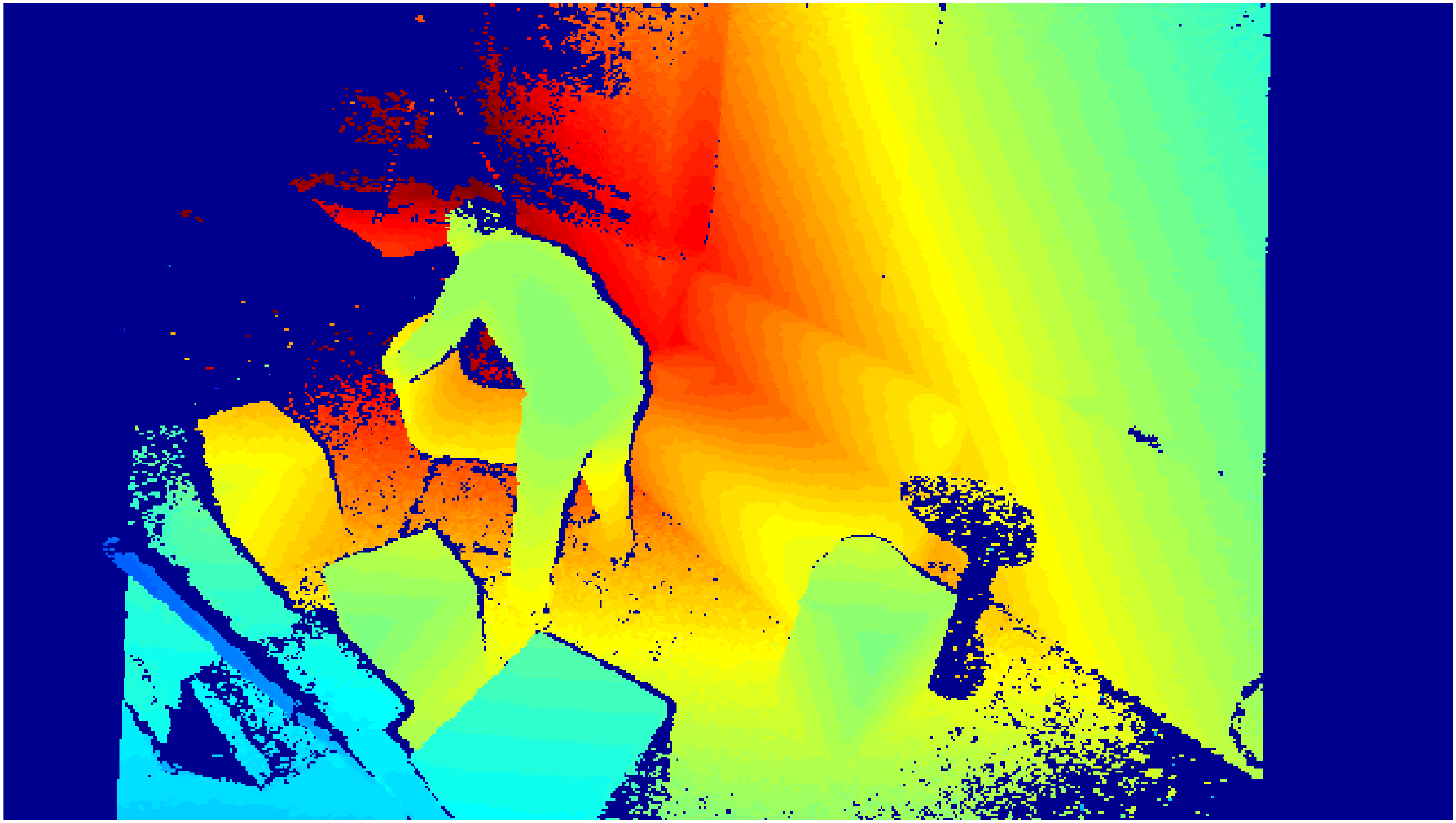}%
		\includegraphics[width=0.125\linewidth]{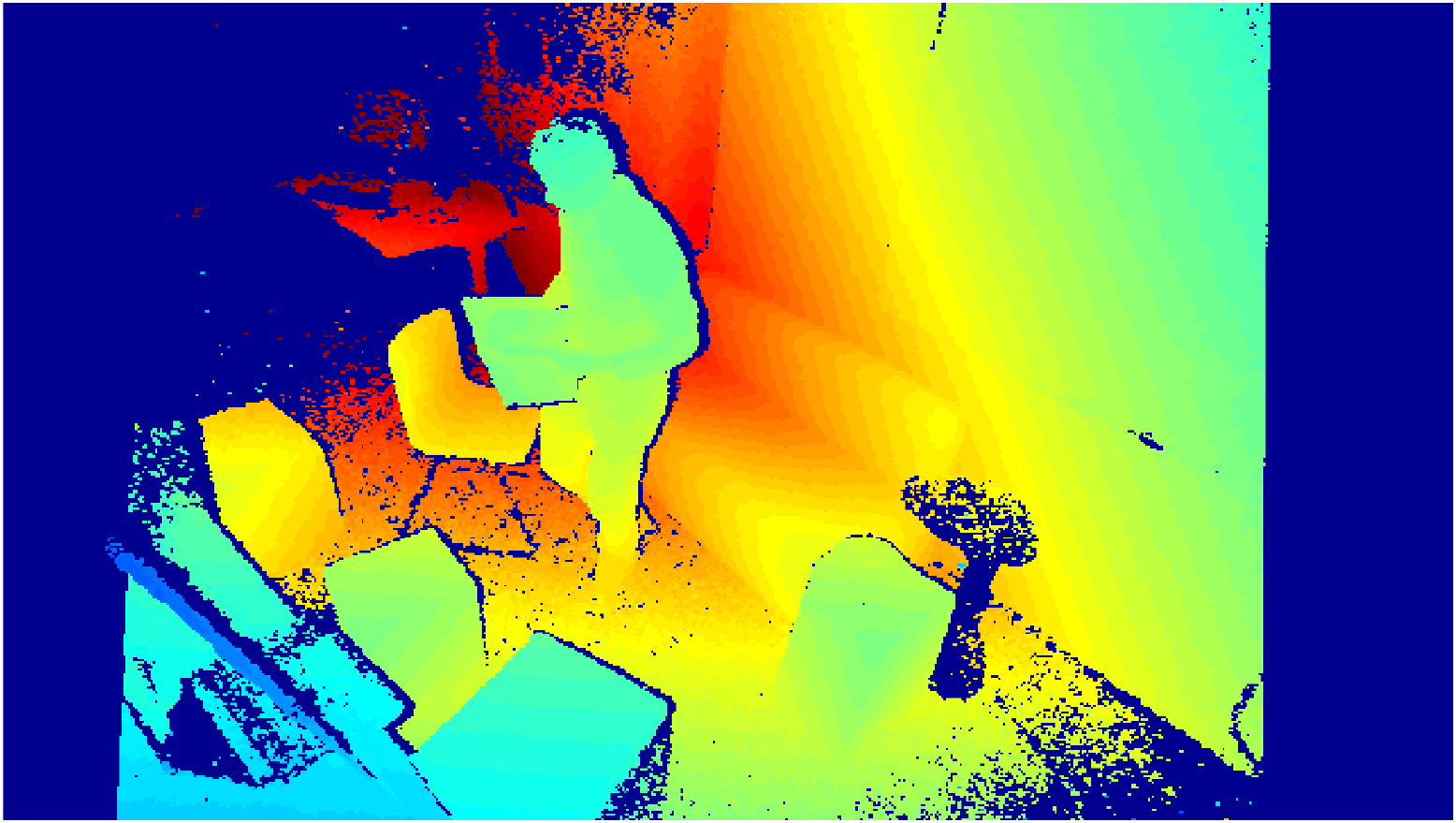}%
		\includegraphics[width=0.125\linewidth]{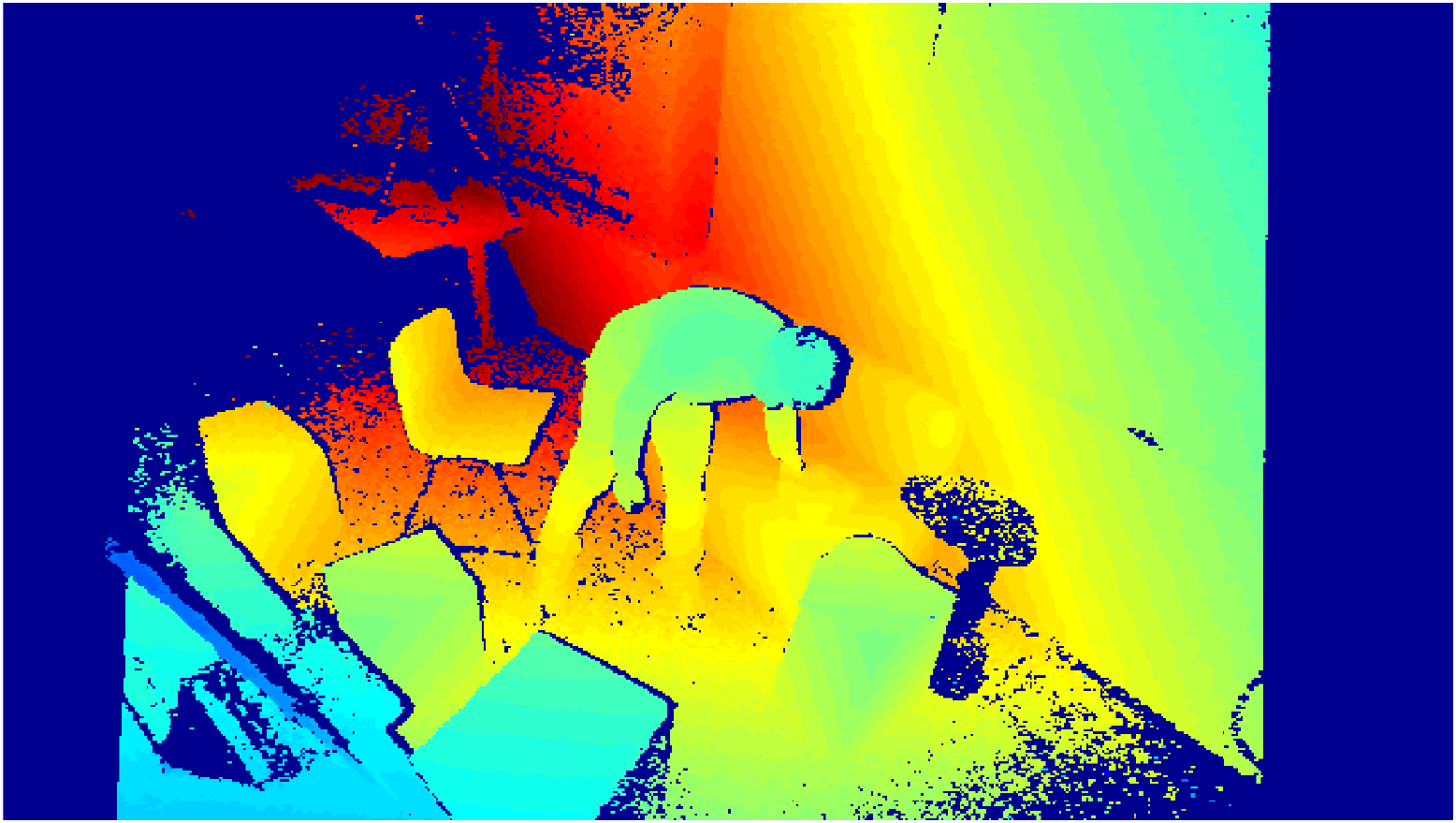}%
		\includegraphics[width=0.125\linewidth]{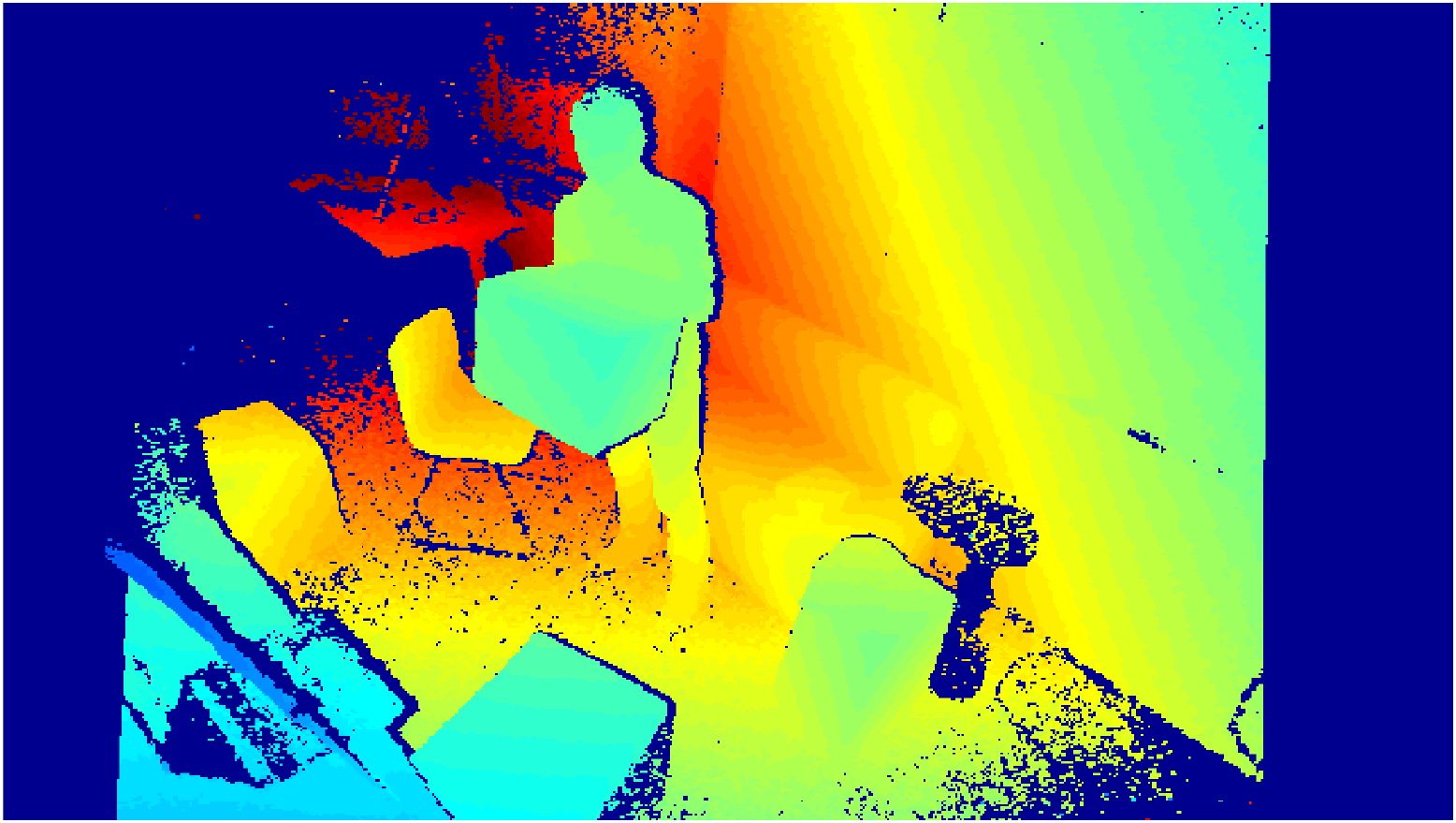}%
		\includegraphics[width=0.125\linewidth]{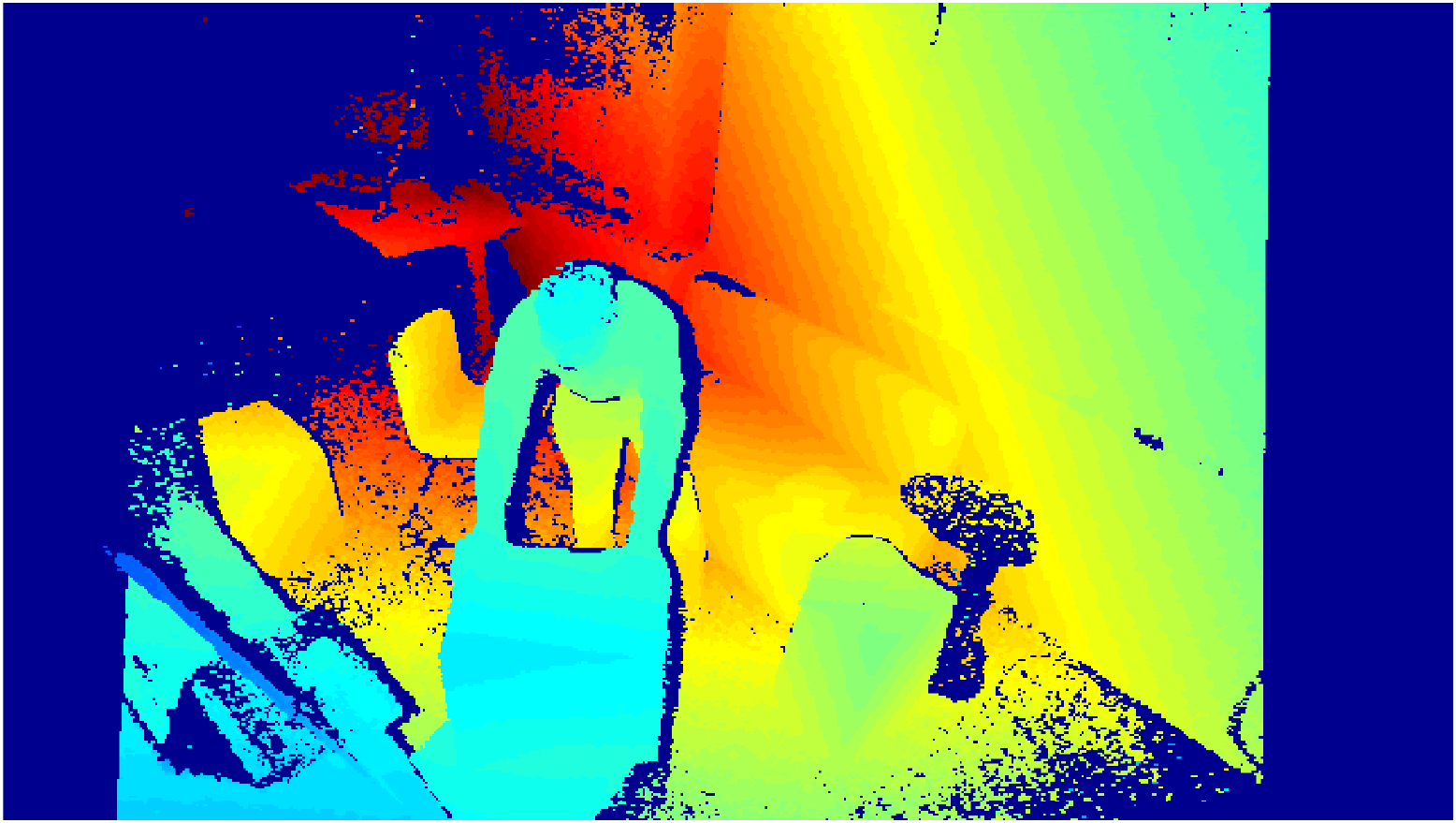}%
		\includegraphics[width=0.125\linewidth]{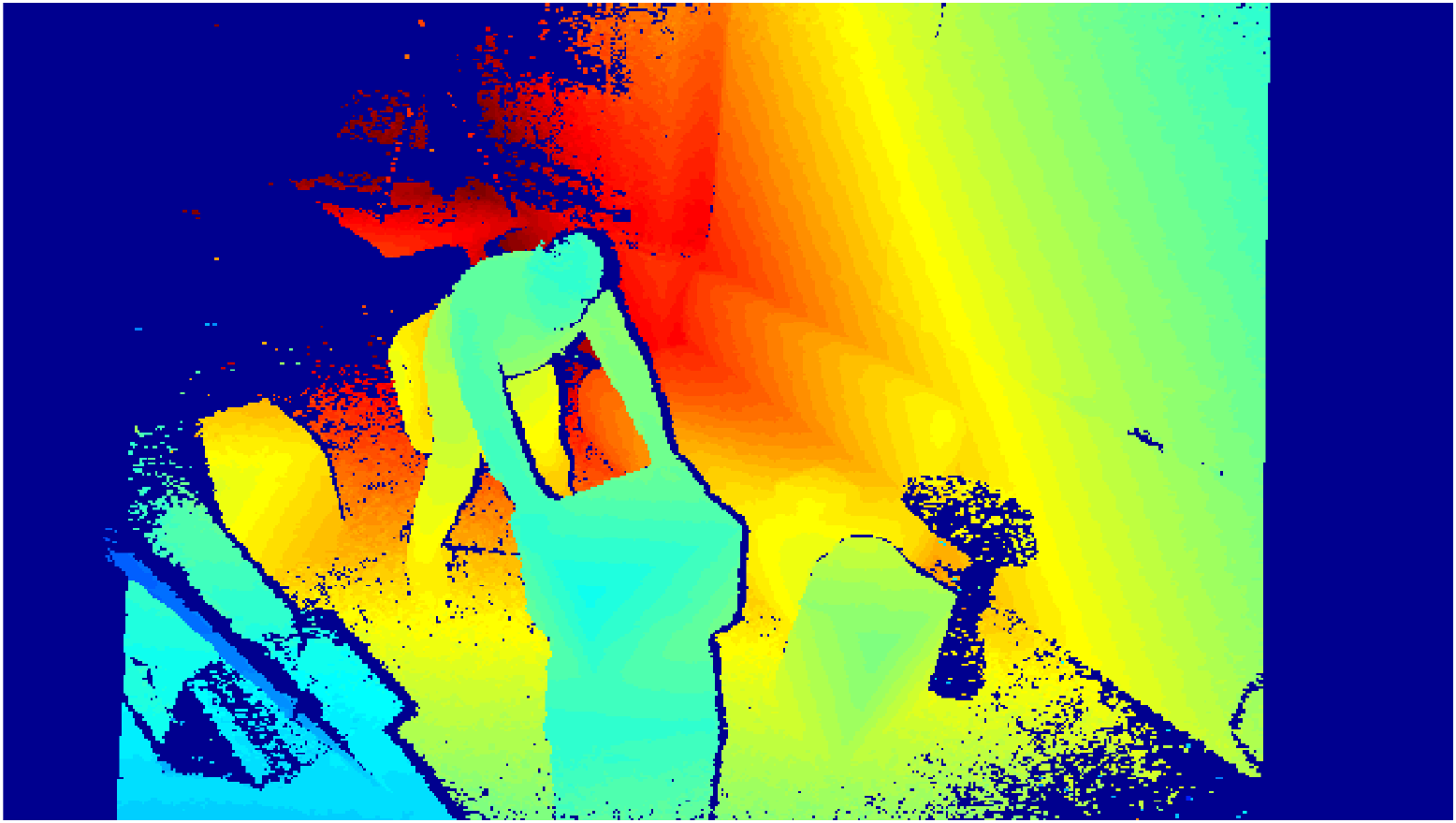}%
		\includegraphics[width=0.125\linewidth]{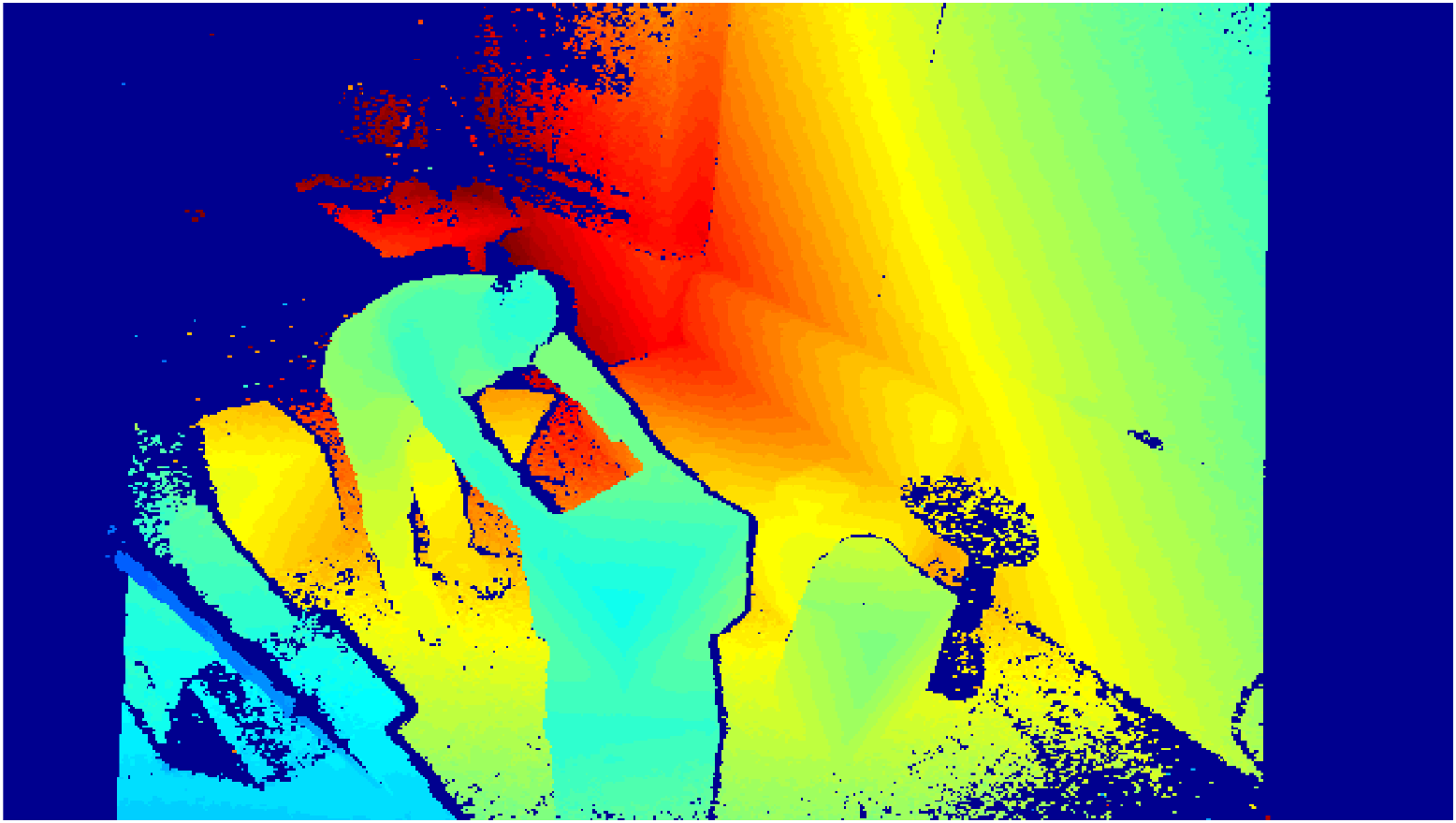}%
		\linebreak
		\includegraphics[width=0.125\linewidth,height=0.15\linewidth]{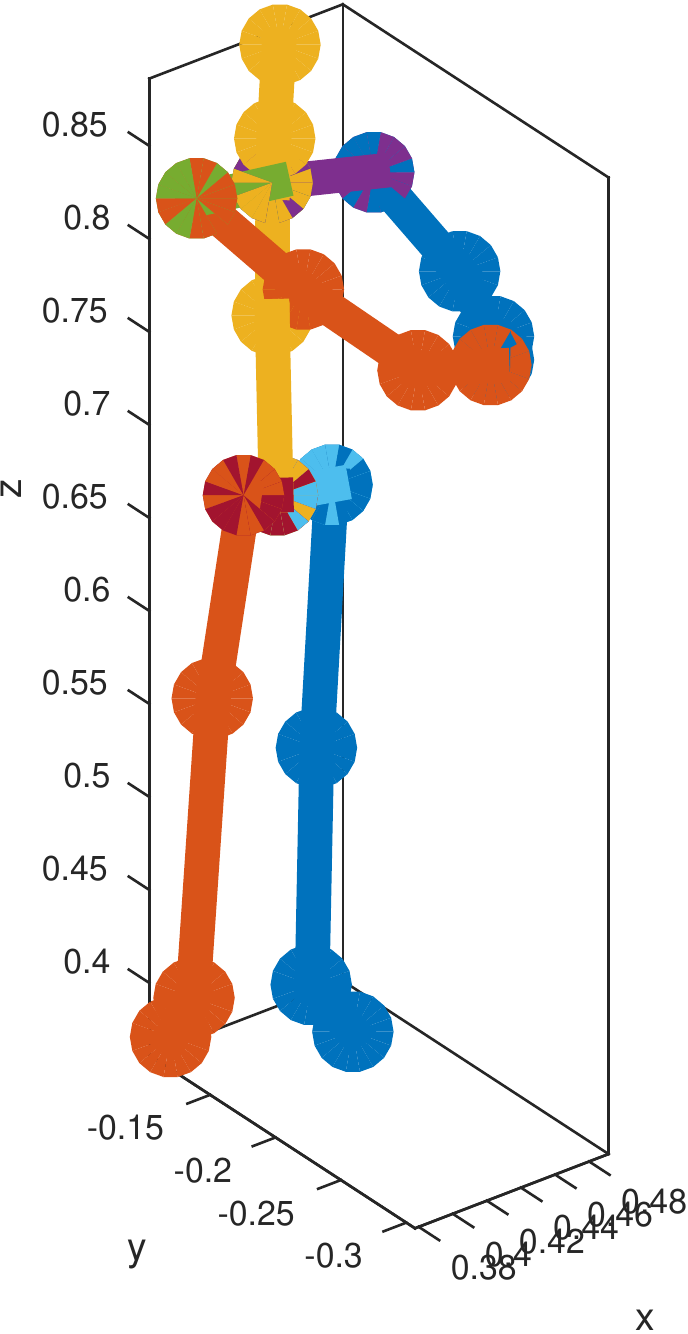}%
		\includegraphics[width=0.125\linewidth,height=0.15\linewidth]{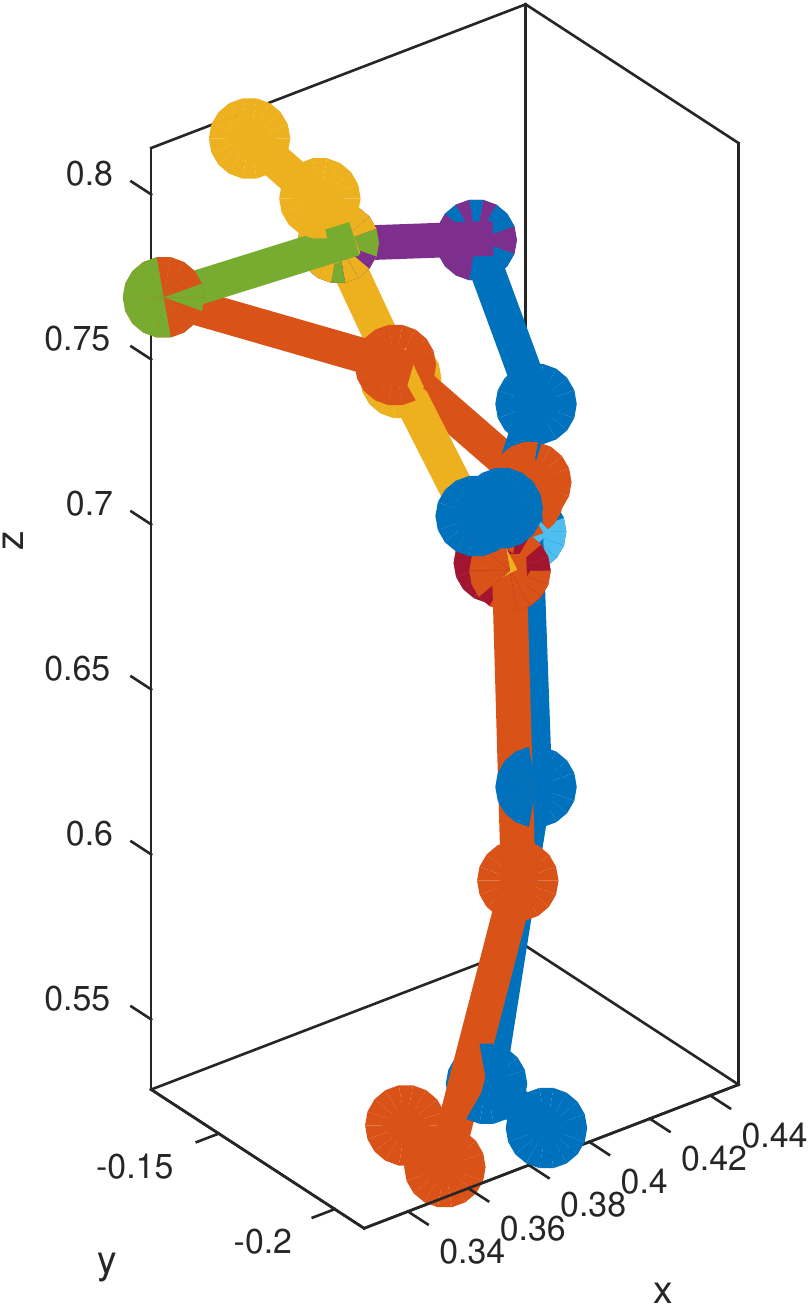}%
		\includegraphics[width=0.125\linewidth,height=0.15\linewidth]{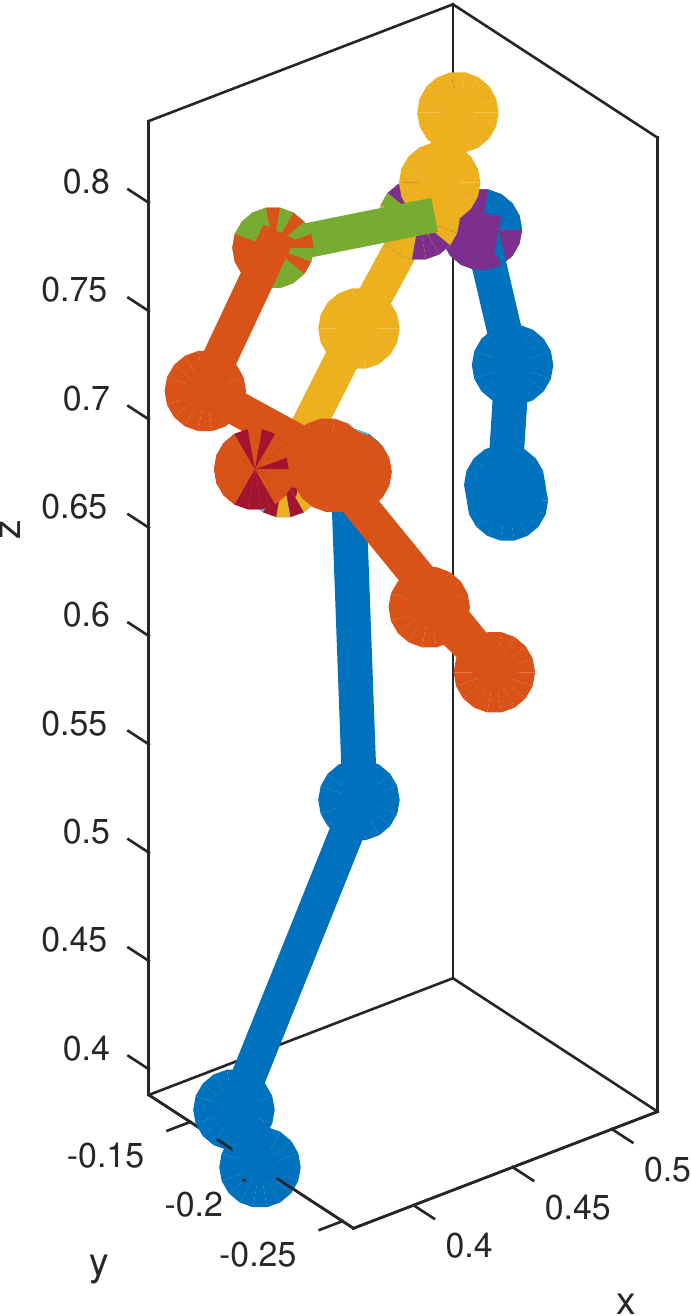}%
		\includegraphics[width=0.125\linewidth,height=0.15\linewidth]{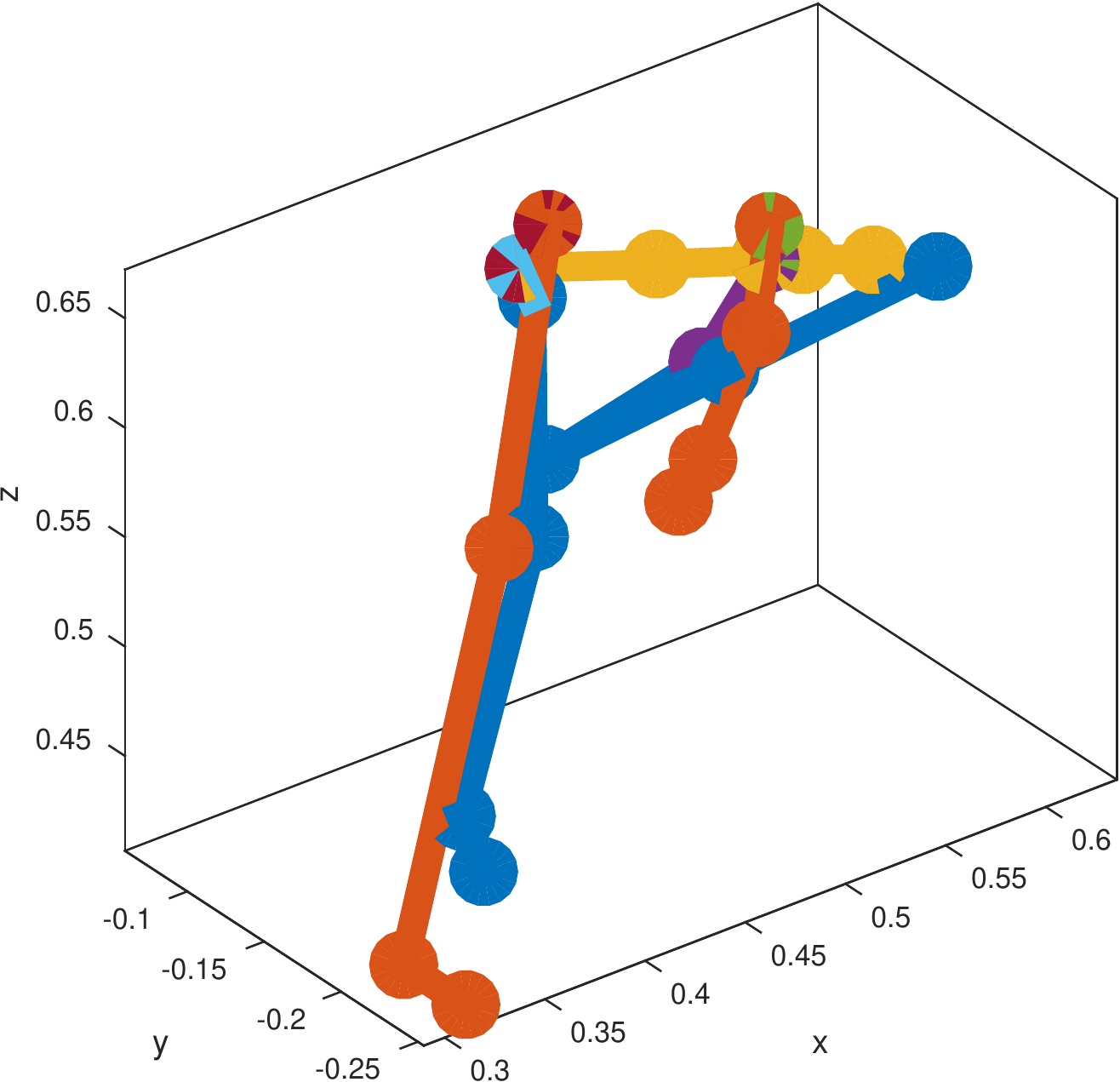}%
		\includegraphics[width=0.125\linewidth,height=0.15\linewidth]{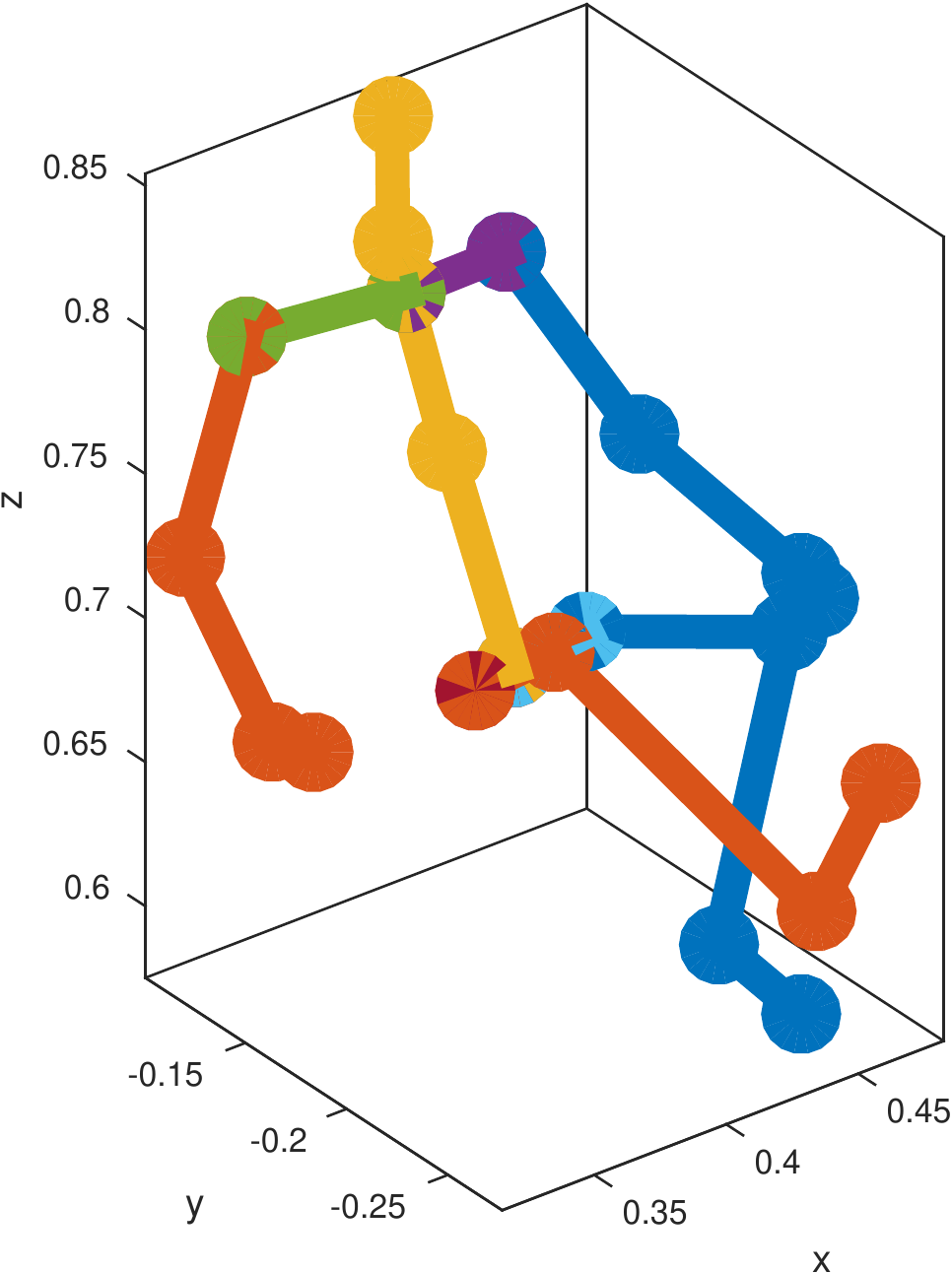}%
		\includegraphics[width=0.125\linewidth,height=0.15\linewidth]{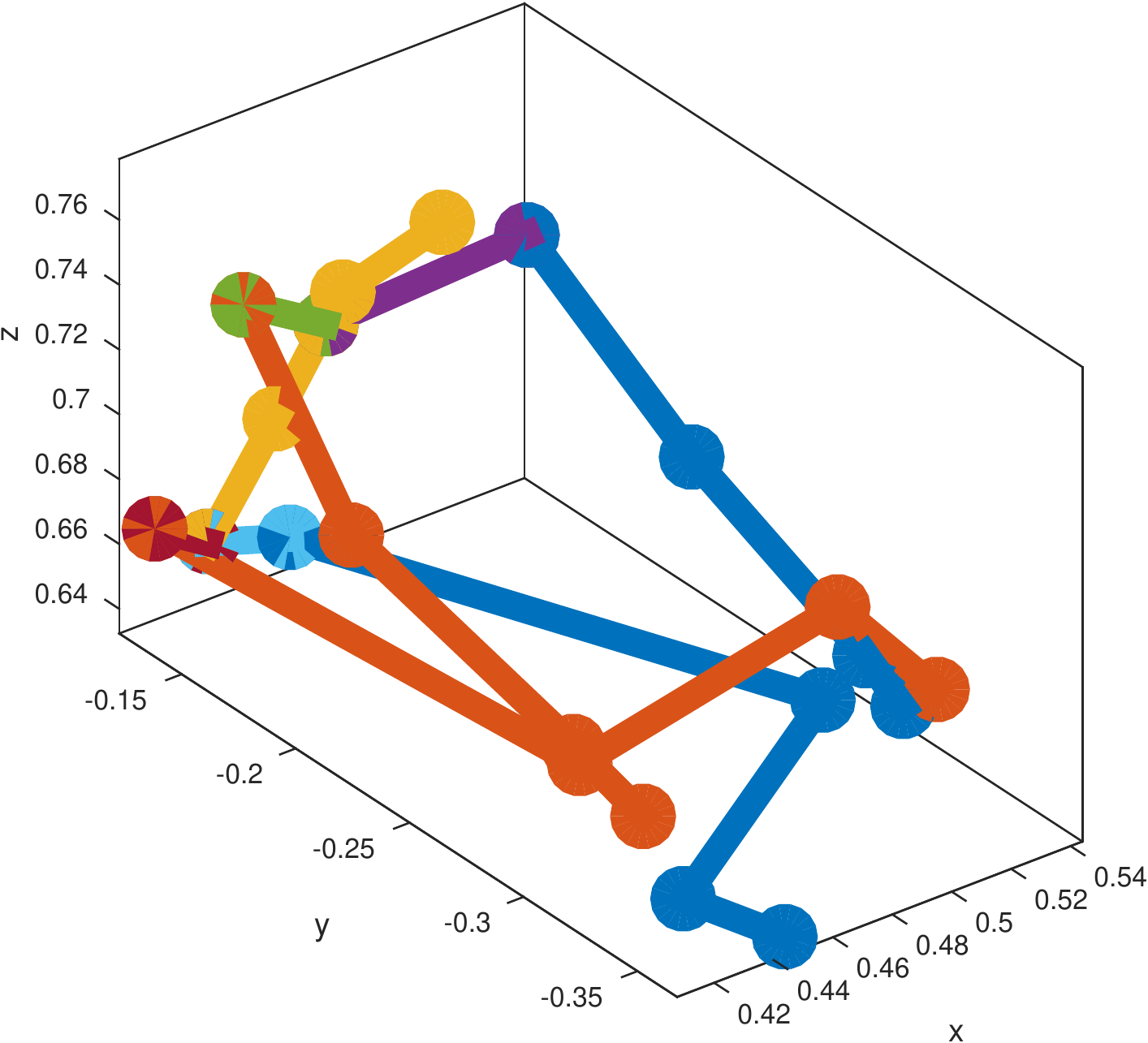}%
		\includegraphics[width=0.125\linewidth,height=0.15\linewidth]{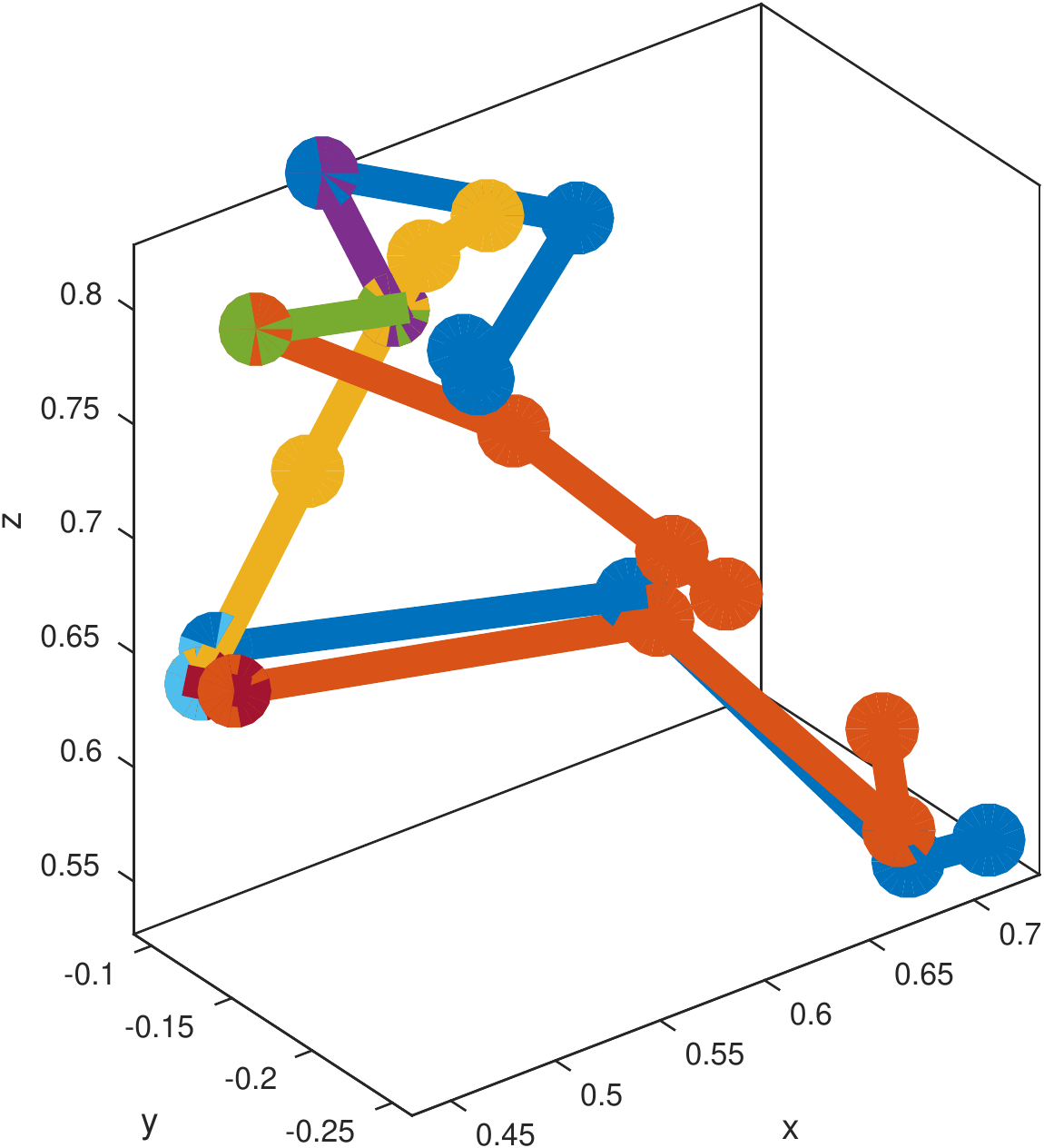}%
		\includegraphics[width=0.125\linewidth,height=0.15\linewidth]{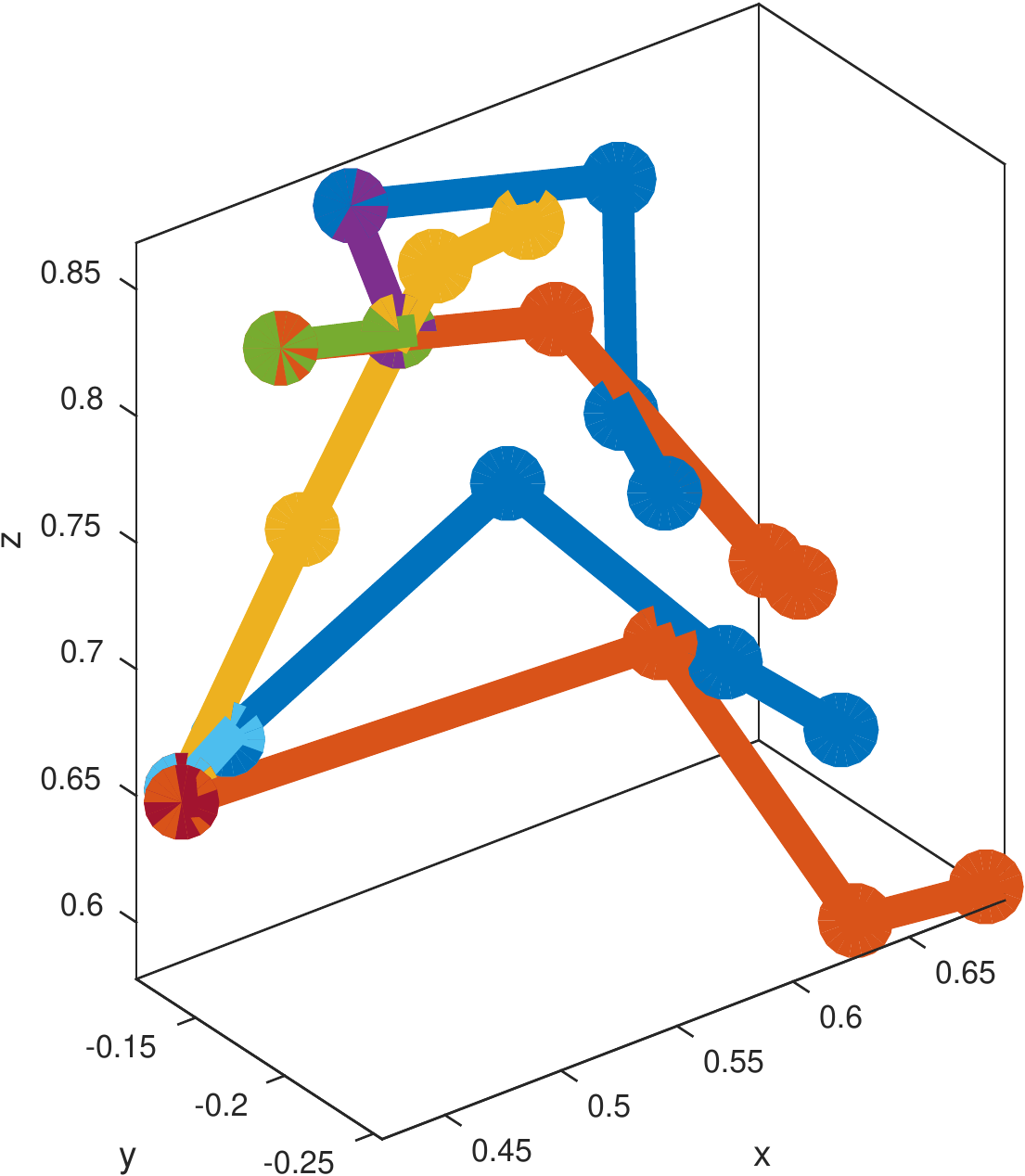}%
		\vspace{-5pt}
		\caption{Kinect V2 skeleton estimation is prone to errors if there is clutter in the scene.
			Row one: Depth images. Row two: Normalized skeletons from Kinect V2.}
		\label{fig:kinectfail}
	\end{figure}
	3D skeletons from the Kinect V2 have been widely used in action recognition. In a clutter free environment, 
	the 3D skeleton estimation from Kinect V2 is
	quite accurate. However, when there is strong clutter in the scene or when there are occlusions, accurate 3D skeleton extraction becomes challenging. 
	Fig.~\ref{fig:kinectfail} shows that the skeleton estimation from Kinect V2 becomes increasingly unreliable as the background clutter increases. 
	When people interact with large objects and their 
	body parts are occluded by these objects, the skeleton estimation fails. 
	
	In addition, the context of an action, \textit{e.g.}, the object being handled and the background objects, is missing in the skeleton representation. 
	The context of an action is important. It is critical when we try to distinguish actions with similar body poses such as reading a book and 
        browsing a cell phone. 
	In this paper, we therefore do not depend on the semantic segmentation and 3D skeleton estimation. Instead, we use the full 4D volume data that 
	contains every bit of information in action recognition. In the following, we show the experimental results on the ground truth data.
	
	\subsection{Ground Truth Experimentation Setup}
	To evaluate the performance of our method we collect a 4D action recognition dataset. We set up four Kinect V2 cameras to capture
	the RGBD images and then we use the real-time solid modeling method to generate 4D volume representation of the dynamic scene. 
	The scene includes not only people
	but also objects such as sofa, tables, chairs, boxes, drawers, cups, and books. 
	There are totally 15 subjects in the dataset. They have different body shapes, gender and 
	heights. The dataset includes 16 actions in the everyday life: drinking, clapping, reading book, calling, 
	playing with phone, bending, squadding, waving hands, sitting, pointing, lifting, opening drawer, pull/pushing, eating, yawning, and kicking. 
	Each action can be done
	in a standing or a sitting pose. Here, action ``sitting'' means sitting without doing anything. Some actions in the dataset can be quite similar
	if we look at the body poses alone. For instance, the poses for reading, playing with phone, and clapping are quite similar.
	We conduct two ground truth tests. Test one involves 14 single-subject videos performed by 14 different subjects. The training is on ten videos and testing is on three videos.
	One video is used for validation. 
	In ground truth test two, we take the trained models from test one and apply them to 
	4 multiple-subject videos, which include 3, 3, 3, and 2 people respectively. 
	The dataset has a total of 68K frames in the training videos,  6K frames in the validating videos, 10K in the single-people testing videos and 6K in 
	the multiple-subject testing videos. 
	We label the videos in a per-frame fashion: each video frame has an action label. We evaluate action recognition 
	using the per-frame accuracy.
	
	We compare our proposed method against different baseline methods. The baselines include: 
	\begin{itemize}
		
		\item \textbf{ShapeContext256 and ShapeContext512}: 3D Shape context is a 3D version of the shape context~\cite{shapecontext} descriptor. 
		The 3D shape context has the height axis and the angle axis uniformly partitioned, and the radial axis logarithmically partitioned.
		We test two versions of the 3D shape context: \textbf{ShapeContext256} has 256 bins and \textbf{ShapeContext512} has 512 bins. 
		We build a deep network whose input 
		is the 3D shape context descriptors. The network uses an LSTM network to aggregate the temporal information.
		
		\item \textbf{Moment}: Moment is another popular shape descriptor. We use the raw moments up to
		order 4. Each element of a moment vector is computed as $\sum_{x,y,z}(x-x_c)^p(y-y_c)^q(z-z_c)^r$, where ($x,y,z$) are the coordinates
		of the occupied voxels and ($x_c,y_c,z_c$) is the volume center. Similar to the above shape context approach, the moment descriptor
		is fed into a CNN for action recognition. 
		
		\item \textbf{Skeletons}: OpenPose \cite{openpose} is one of the state-of-the-art stick figure detectors on RGB images. It is able to 
		give reasonably good pose estimation results on each color video frame. We normalize the positions of the joints of each subject using the
		neck point and then concatenate the $xy$ coordinates from all four cameras into a feature vector. We train a deep network using similar
		approach to the above shape context method.
		
		\item \textbf{Color+Depth}: This follows the scheme of standard action recognition methods on 2D images. 
		In this method, we find the bounding boxes of each person based on our tracking result. We crop the color and depth images 
		of each person in the video from all the cameras. The cropped color and depth video are used in action recognition. We train a deep neural 
		network using the cropped color and depth images and their action labels. To be fair, we do not use motion
		in all the methods in this paper.  
		
		\item \textbf{PointNet}: PointNet \cite{pointnet} is one of state-of-the-art deep learning methods for object recognition and semantic segmentation 
		on 3D point clouds. We extend the PointNet model to include an LSTM layer so that it can handle sequential data for action recognition. The network 
		can be trained end-to-end using the point clouds from the four RGBD images. 
		
		\item \textbf{Variations of the Proposed Method}:
		We also compare the proposed method with a few variations of the proposed method. 
		In the first variation, we throw away the internal voxels in the volume data 
		but only keep the shell of each object. The shells can be directly computed from the merged 3D point cloud.
		With the shell representation, we train a deep network with exactly the same structure as our proposed method.
		We denote this method as the shell-volume approach and the method using solid model as the solid-volume approach. 
		To verify the effectiveness of different structures in our deep neural network, we also perturb the network structures. We test three versions.
		In \textbf{Conv3D+LSTM+G}, we remove the attention network.
		In \textbf{Conv3D+LSTM+ATT}, we remove the global max-pooling.
		Action4D-Net uses the full model \textbf{Conv3D+LSTM+ATT+G}.
		
	\end{itemize}
	
	All the models are implemented using PyTorch and trained from scratch. We use the same training, testing and validating splits when evaluating different
	methods. 
	The validating set is used to determine the stopping point and model selection. 
	The CNN architectures we used for \textbf{ShapeContext256}, \textbf{ShapeContext512}, 
	\textbf{Moment}, \textbf{Skeleton}, and \textbf{Color+Depth} can all
	achieve almost $100\%$ accuracy on the training dataset, which suggests 
	that the capacities of the proposed models are enough for the action recognition tasks.
	
	\subsection{Ground Truth Experimentation Results}
\begin{figure}[tb]
		\centering
		\subfigure[]{\includegraphics[width=0.5\linewidth]{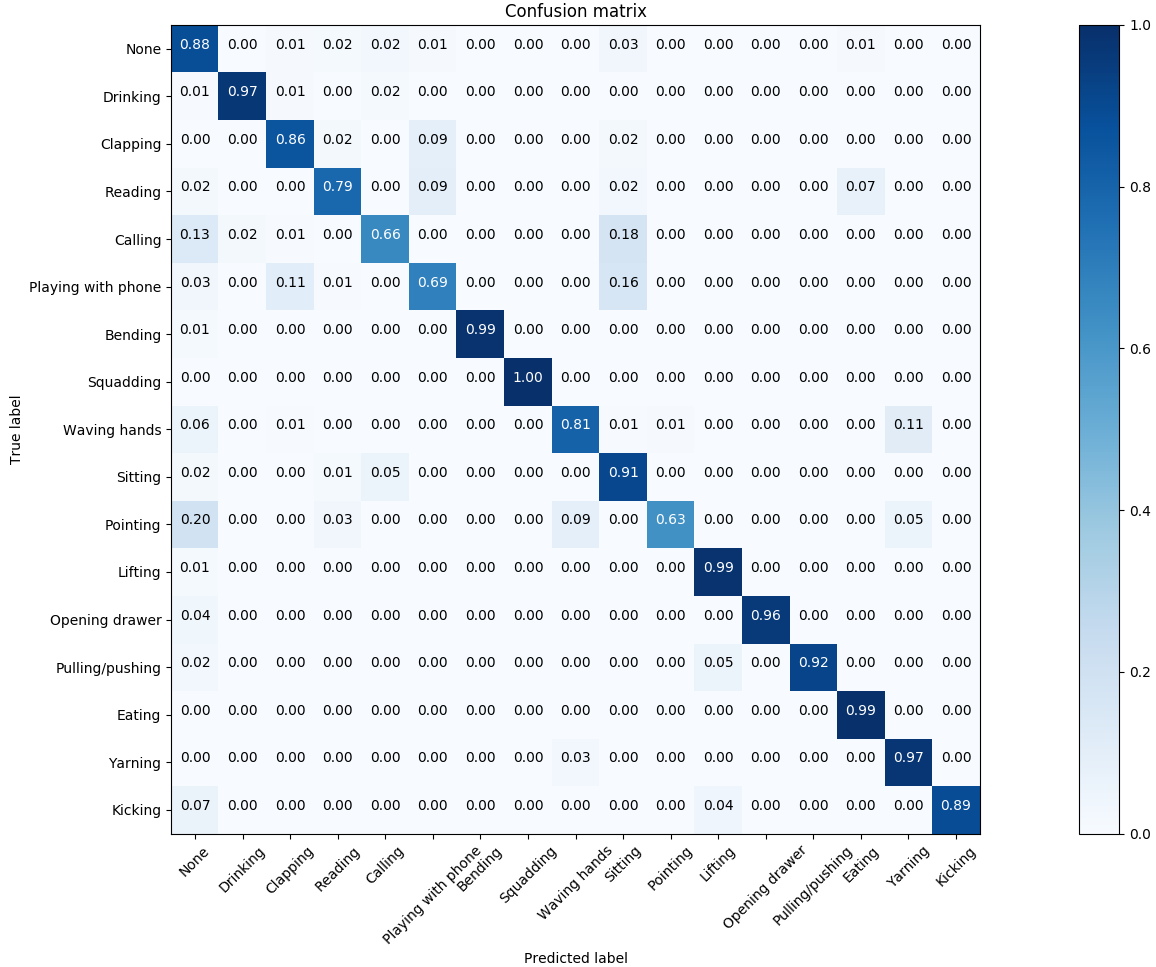}}%
		\subfigure[]{\includegraphics[width=0.5\linewidth]{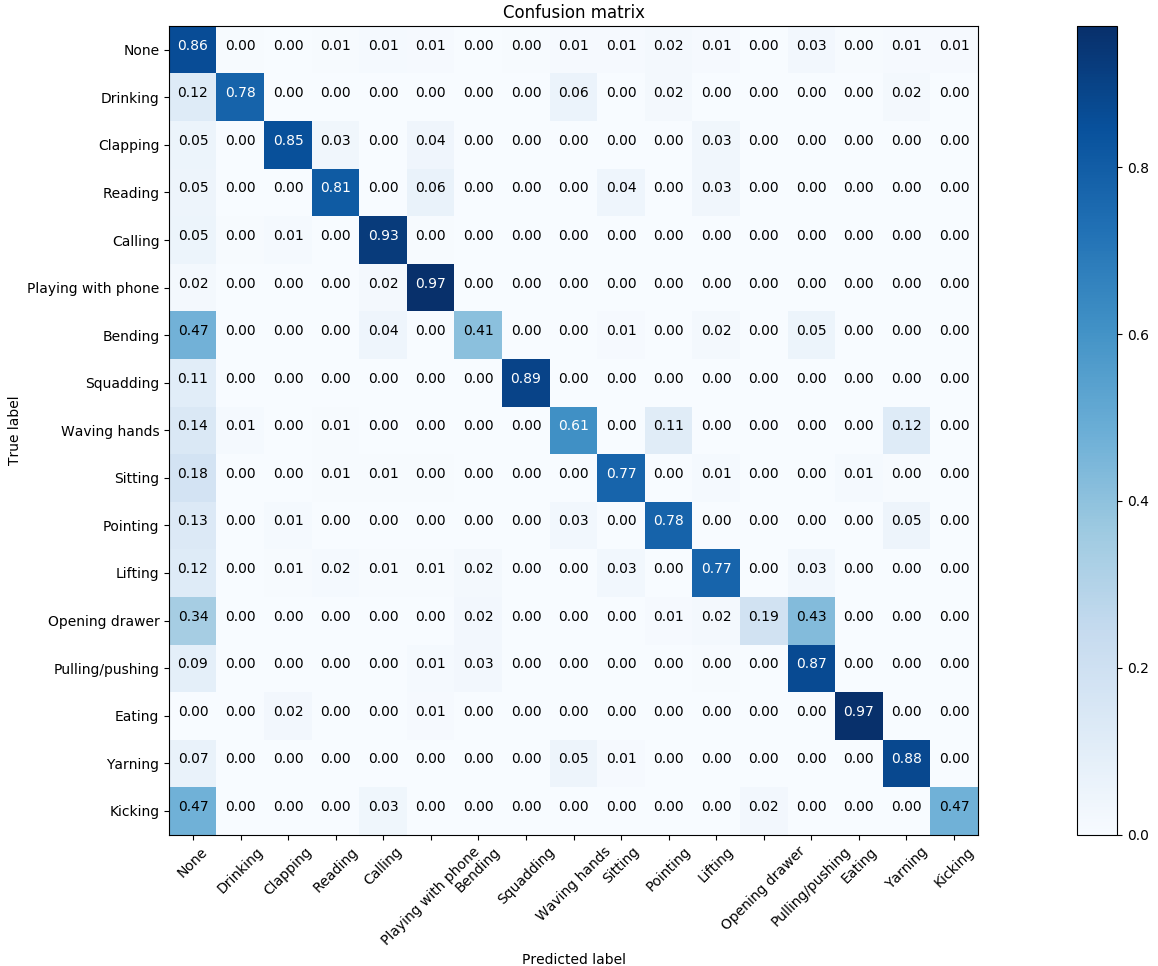}}%
		\caption{Confusion matrices for ground truth test one in (a) and two in (b).}
		\label{fig:confuse}
	\end{figure}

	We report the action recognition results in Table~\ref{tab:testing} and Table~\ref{tab:testing:mp}. 
        Please view the action recognition result videos at \url{http://www.hao-jiang.net/projects/action4d/video.html}. 
	We report two accuracy numbers for each test. Acc is more strict; we deem action recognition is correct if and only if the action prediction result
	matches the ground truth label of the corresponding video frame. 
	One issue about this criterion is that at the action boundaries accurate labeling is hard. For transient actions, 
	the small offset of labeling may cause the mismatch between the detection result and the ground truth. To reduce this problem, 
	we define another accuracy, 
	revised accuracy (RAcc). For RAcc, an action classification is correct if and only if the predicted action label is the same as the ground truth label 
	of a frame within the window of plus/minus
	three relative to the current video frame. We found RAcc is more consistent with the visual inspection result.
	As shown in Table~\ref{tab:testing} and Table~\ref{tab:testing:mp}, our proposed method gives the highest action recognition accuracy among all the competing methods.
	The confusion matrices in Fig.~\ref{fig:confuse} confirm that our proposed method gives accurate result.
	
	\begin{table*}[!ht]
		\centering
		\caption{Evaluation of the proposed models and several baselines on the testing split of the dataset in ground truth test one, 
which consists of three people. We show both the accuracies (Acc) and the revised accuracies (RAcc) of all the evaluated models. The numbers show the percentages.}
		\label{tab:testing}
		\resizebox{0.9\textwidth}{!}{%
			\begin{tabular}{l*{8}{c}}
				\hline
				\multirow{2}{*}{Models} & \multicolumn{2}{c}{Person 1} & \multicolumn{2}{c}{Person 2} & \multicolumn{2}{c}{Person 3}  & \multicolumn{2}{c}{Average} \\ 
				\cline{2-9} 
				& Acc & RAcc  & Acc & RAcc & Acc & RAcc & Acc & RAcc \\ \hline
				ShapeContxt256	&	56.9	&	63.1	&	51.5	&	56.5	&	55.7	&	61.6	&	54.7	&	60.5	\\
				ShapeContxt512	&	55.2	&	60.5	&	47.6	&	53.1	&	56.8	&	62.6	&	53.4	&	58.9	\\
				Moments	&	37.4	&	44.9	&	44.9	&	54	&	38.4	&	47.1	&	40.1	&	48.6	\\
				Color+Depth	&	53.6	&	60.5	&	64.1	&	71.7	&	52.7	&	60.1	&	56.6	&	63.9	\\
				Skeleton	&	60.1	&	66.0	&	55.4	&	62.0	&	64.0	&	69.9	&	60.2	&	66.1	\\
				PointNet	&	38.6	&	46.3	&	44.0	&	53.1	&	40.1	&	49.3	&	41.1	&	49.5	\\
				\hline
				Solid Volume \\
				\hline
				Our Conv3D+LSTM+G	&	75.7	&	82.1	&	73.4	&	81.5	&	73.2	&	80.4	&	74.1	&	81.3	\\
				Our Conv3D+LSTM+ATT	&	80.4	&	86.4	&	78.8	&	86.7	&	75.7	&	82.8	&	78.3	&	85.2	\\
				Our Conv3d+LSTM+ATT+G	&	80.7	&	87.0	&	79.6	&	87.3	&	\textbf{77.8}	&	\textbf{84.8}	&	79.4	&	86.3	\\
				\hline
				Shell Volume \\
				\hline
				Our Conv3D+LSTM+G	&	72.6	&	78.3	&	71.8	&	79.6	&	69.2	&	76.3	&	71.1	&	78.0	\\
				Our Conv3D+LSTM+ATT	&	74.7	&	81.0	&	75.4	&	82.9	&	74.5	&	81.4	&	74.8	&	81.7	\\
				Our Conv3d+LSTM+ATT+G	&	\textbf{81.2}	&	\textbf{87.5}	&	\textbf{83.8}	&	\textbf{91.1}	&	74.6	&	81.6	&	\textbf{79.7}	&	\textbf{86.6}\\
				\hline
		\end{tabular}}
	\end{table*}
	\normalsize

	Table~\ref{tab:testing} shows the accuracies of different competing methods in ground truth test one.
	In this test, our proposed methods achieve the highest average revised accuracy (RAcc) $86.6\%$ and $86.3\%$ for the shell and the solid
	model respectively. Our method's accuracy improves by more than $30\%$ over the competing methods such as \textbf{ShapeContext}, 
	\textbf{Moment}, \textbf{Color+Depth}, \textbf{Skeleton} and \textbf{PointNet}. We also achieve the highest accuracies in each individual
	test. 
	
	These results are not a surprise. The handcrafted features such as shape context and moments are not as strong as the 
	learned features from deep learning especially when there is strong background clutter. 
	Fine-tuning the handcrafted features is often not helpful; as shown in \tablename~\ref{tab:testing}, the shape context method
	with 512 bins is in fact worse than the one with 256 bins.  
	The \textbf{PointNet}
	gives low accuracies in this experiment. This is likely due to the strong clutter and because \textbf{PointNet} has to
	sample the point clouds to fit into the GPU memory. The \textbf{Color+Depth} and \textbf{Skeleton} approaches
	perform better than other handcrafted feature methods, but they give much worse results than our proposed method. Both \textbf{Color+Depth} and \textbf{Skeleton}
	methods
	are also dependent on the camera views: if the camera settings are different, we have to retrain the model. In contrast, our
	proposed method can be used in different camera settings without retraining.    
	
	In the ground truth test two, each video involves multiple people.
	As shown in Table~\ref{tab:testing:mp}, our method still gives the highest accuracy among all the methods.
	\textbf{Color+Depth} gives reasonable good result but the accuracy is still at least $20\%$ lower than our method.
	\textbf{Skeleton} gives worse result than ours due to the multiple people mutual occlusions.
	Our method appears to give higher accuracy in the multiple-subject test than in the single person test.
	This is mostly because some testing groups include one person who also appears in the training dataset.
	In ground truth test two, group 2 does not include any subjects from the training dataset; its result looks consistent with the result in 
	Table~\ref{tab:testing}.      
	
	To justify why we use the attention plus global network structure, we compare the results with or without each component.
	As shown in Table~\ref{tab:testing} and Table~\ref{tab:testing:mp}, our proposed network with both of the components gives the best results.
	The attention and global sub-network are indeed helpful.
	
	An interesting phenomenon is that the shell-volume approach seems give slightly better result in ground truth test one. This is indeed a
	bit surprise since the solid model does provide more information about the inside and outside part of each object.
	However, in Table~\ref{tab:testing:mp} in which we evaluate our method on the more cluttered multiple-subject videos, the solid model representation gives much higher accuracy. 
	This shows the solid-volume approach is more resistant to the clutter than the shell-volume approach.
	\begin{table*}[!ht]
		\centering
		\caption{Evaluation of the proposed models and several baselines in ground truth test two, which involves multiple people. }
		\label{tab:testing:mp}
		\resizebox{0.9\textwidth}{!}{%
			\begin{tabular}{l*{10}{c}}
				\hline
				\multirow{2}{*}{Models} & \multicolumn{2}{c}{Group 1} & \multicolumn{2}{c}{Group 2} & \multicolumn{2}{c}{Group 3}  & \multicolumn{2}{c}{Group 4} &\multicolumn{2}{c}{Average} \\ 
				\cline{2-11} 
				& Acc & RAcc  & Acc & RAcc & Acc & RAcc & Acc & RAcc & Acc & RAcc \\ \hline
				ShapeContxt	&	31.9	&	37.6	&	32.3	&	37.8	&	32.6	&	39.6	&	58.8	&	65.2	&	37.5	&	43.6	\\
				ShapeContxt16	&	31.5	&	36.2	&	23.7	&	28.3	&	30.1	&	35.3	&	57.5	&	63	&	34.2	&	39.1	\\
				Moments	&	33.4	&	41.2	&	34.7	&	44	&	37.8	&	46.6	&	40.7	&	48.2	&	36.2	&	44.5	\\
				Color+Depth	&	41.2	&	48.2	&	46.9	&	55.4	&	54	&	63.7	&	64.1	&	70.9	&	46.1	&	56.6	\\
				Skeleton	&	44.9	&	52.7	&	44.9	&	51.2	&	52.1	&	56.6	&	49.7	&	56.1	&	50	&	58.1	\\
				PointNet	&	45.4	&	55.8	&	49.8	&	60.4	&	50.1	&	62	&	47.4	&	57	&	47.2	&	53.7	\\
				\hline
				Solid Volume \\
				\hline
				Conv3D+LSTM+G	&	64.4	&	72.9	&	62.7	&	70.5	&	69.5	&	77.8	&	87.8	&	95.4	&	69.7	&	77.5	\\
				Conv3D+LSTM+ATT	&	72.4	&	80.6	&	71.5	&	79.5	&	77.7	&	86.4	&	84.8	&	92.4	&	75.7	&	83.4	\\
				Conv3d+LSTM+ATT+G	&		\textbf{76.2}	&		\textbf{83.6}	&		\textbf{76}	&		\textbf{83.8}	&		\textbf{84.7}	&		\textbf{92.6}	&		\textbf{89.5}	&		\textbf{96.6}	&		\textbf{80.6}	&		\textbf{88.3}	\\
				\hline
				Shell Volume \\
				\hline
				Conv3D+LSTM+G	&	64.6	&	71.6	&	66.1	&	73.8	&	69.2	&	78.7	&	82.4	&	89.4	&	69.5	&	77.4	\\
				Conv3D+LSTM+ATT	&	67.5	&	75.2	&	67.5	&	75.7	&	80.1	&	87.8	&	81.8	&	89.2	&	73.2	&	81	\\
				Conv3d+LSTM+ATT+G	&	72	&	78.6	&	71.8	&	79.7	&	84.3	&	91.8	&	84.6	&	91.4	&	77.2	&	84.4	\\
				\hline
		\end{tabular}}
	\end{table*}
	
	Table~\ref{tab:testing} and Table~\ref{tab:testing:mp} show that our proposed method consistently gives much better result than all the competing methods.
	The high accuracy also benefits from our reliable people tracker, which give $100\%$ tracking rate for all the testing and training videos.  
	Our method is also fast, with a single GTX1080 TI, our method is able to track 10 people and infer their actions at 15 frames per second.
	
	Our method still confuses some actions as seen in the confusion matrices. This is mostly due to the sometimes very noisy data from the Kinect sensor
	and that causes the losing details. 
	Using better depth cameras and better time
	synchronization, our action recognition results can be further improved. Moreover, we can further include other voxel attributes such as color and use 
	multi-resolution volume data to achieve more robust results.   
	
	\section{Conclusion}  
	We propose a novel 4D action recognition method, the Action4D, which is able to generate 4D solid model of the environment, track each 
	person in the volume and infer the actions of each subject in real time. Our method is able to handle multiple people and strong clutter.
	Our experimental results confirm that our method gives the best performance among different competing methods. 
	The proposed method can be deployed to enable different applications to enhance how people interact with the environment.


\begin{thebibliography}{50}

\bibitem{kidsroom} KidsRomm, \url{http://vismod.media.mit.edu/vismod/demos/kidsroom/kidsroom.html}.

\bibitem{greg} Wang, Y., Jiang, H., Drew, M.S., Li, Z.N. and Mori, G.: Unsupervised Discovery of Action Classes. CVPR 2006.

\bibitem{efros} Efros, A.A., Berg, A.C., Mori, G. and Malik, J.: Recognizing Action at A Distance. ICCV 2003.

\bibitem{kth} Laptev, I. and Lindeberg, T.: Space-Time Interest Points. ICCV 2003.

\bibitem{irani1} Blank, M., Gorelick, L., Shechtman, E., Irani, M. and Basri, R.: Actions as Space-Time Shapes.  ICCV 2005.

\bibitem{irani2} Shechtman, E.  and Irani, M.: Space-Time Behavior Based Correlation.  CVPR 2005.

\bibitem{actiontube} Kalogeiton, V., Weinzaepfel, P., Ferrari, V., Schmid, C.: Action Tubelet Detector for Spatio-Temporal Action Localization. ICCV 2017.

\bibitem{threestream} Zolfaghari, M., Oliveira, G.L., Sedaghat, N., Brox, T.: Chained Multi-stream Networks Exploiting Pose, Motion, and Appearance for Action Classification and Detection. ICCV 2017.

\bibitem{anothericcv17} Singh, G., Saha, S., Sapienza, M., Torr, P., Cuzzolin F.: Online Real-time Multiple Spatiotemporal Action Localisation and Prediction. ICCV 2017.

\bibitem{eccv16joints} 
Li, Y., Lan, C., Xing, J., Zeng, W., Yuan, C. Liu, J.: Online Human Action Detection using Joint Classification-Regression Recurrent Neural Networks. ECCV 2016.

\bibitem{3dnetwork} Diba, A., Sharma, V., Gool, L.V.: Deep Temporal Linear Encoding Network. CVPR 2017. 

\bibitem{ntu} Shahroudy, A., Liu, J., Ng, T.T. and Wang, G.: NTU RGB+D: A Large Scale Dataset for 3D Human Activity Analysis. CVPR 2016.

\bibitem{depth1}  Wang, P., Li, W., Gao, Z., Zhang, Y., Tang C., Ogunbona, P.: Scene Flow to Action Map: A New Representation for RGB-D based
Action Recognition with Convolutional Neural Networks. 
CVPR 2017.

\bibitem{cornellrgbd} 
Sung, J., Ponce, C., Selman, B., Saxena, A.: Unstructured human activity detection from RGBD images. ICRA 2012.

\bibitem{utaustin} Xia, L., Chen, C.C., Aggarwal, J.K.: View Invariant Human Action Recognition Using Histograms of 3D Joints. CVPR Workshop 2012.

\bibitem{visualhull}
Laurentini, A.: The Visual Hull Concept for Silhouette-Based Image Understanding,
TPAMI, Volume 16 Issue 2, February 1994. 

\bibitem{vh1}
Canton-Ferrer, C., Casas, J.R., Pard\`{a}s, M.: Human model and motion based 3D action recognition in multiple view scenarios. European Signal Processing 2006.

\bibitem{vh2}
Weinland, D., Ronfard, R. and Boyer. E.: Free viewpoint action recognition using motion history
volumes. CVIU, 104(2):249–257, 2006.

\bibitem{vh3}
 	Holte, 	M.B.,
	Tran, C.,
Trivedi, M.M., 
	Moeslund, T.B.:
Human action recognition using multiple views: a comparative perspective on recent developments.
J-HGBU '11 Proceedings of the 2011 joint ACM workshop on Human gesture and behavior understanding
Pages 47-52. 

\bibitem{fusion4d}
  Dou, M., 	
	Khamis, S.,
	Degtyarev, Y.,
	Davidson, P.,
	Fanello, S.R.,
	Kowdle, A.,
	Escolano, S.O.,
	 Rhemann, R.,
	 Kim, D.,
	 Taylor, J.,
	 Kohli, P.,
	 Tankovich, V.,
	 Izadi, S.: Fusion4D: real-time performance capture of challenging scenes. ACM SIGGRAPH 2016.

\bibitem{skeleton1}
Lie, J., Wang, G., Hu, P., Duan, L.Y., Kot, A.C.: Global Context-Aware Attention LSTM Networks for 3D Action Recognition.
CVPR 2017.

\bibitem{skeleton2}
 Rahmani, H. and  Bennamoun, M.: Learning Action Recognition Model From Depth and Skeleton Videos.
ICCV 2017.

\bibitem{skeleton3}
 Wang, H.,
 Wang, L.: Modeling Temporal Dynamics and Spatial Configurations of Actions Using
Two-Stream Recurrent Neural Networks.
CVPR 2017.

\bibitem{skeleton4}
Lee, I., Kim, D., Kang, S., Lee,S.: Ensemble Deep Learning for Skeleton-based Action Recognition
using Temporal Sliding LSTM networks.
ICCV 2017.

\bibitem{skeleton5}
 Zhang, P.,  Lan, C., Xing, J., Zeng, W., Xue, J., Zheng, N.: View Adaptive Recurrent Neural Networks for High Performance Human Action Recognition from Skeleton Data.
ICCV 2017.


\bibitem{skeleton7}
Huang, Z., Wan, C., Probst, T., Gool, L.V.: Deep Learning on Lie Groups for Skeleton-based Action Recognition. CVPR 2017.

\bibitem{shapecontext} 
Belongie, S., Malik J. and Puzicha, J.: Shape Matching and Object Recognition Using Shape Contexts.
TPAMI 2002.

\bibitem{openpose}
 Cao, Z., Simon, T., Wei, S.E. and Sheikh, Y.: Realtime Multi-Person 2D Pose Estimation using Part Affinity Fields.
CVPR 2017.

\bibitem{pointnet} Qi, C.R.,  Su, H.,  Mo, K.,   Guibas, L.J.: PointNet: Deep Learning on Point Sets for 3D Classification and Segmentation.
CVPR 2017.

\bibitem{buffy}
Ferrari, V., Marin-Jimenez, M. and Zisserman, A.:
Progressive Search Space Reduction for Human Pose Estimation.
CVPR 2008.

\bibitem{fcn}
Shelhamer, E., Long, J., Darrell, T.: Fully Convolutional Models for Semantic Segmentation.
PAMI 2016.

\bibitem{bahdanau2014neural} Bahdanau, D.,  Cho, K. and Bengio, Y.: Neural machine translation by jointly learning to align and translate. arXiv preprint arXiv:1409.0473 2014.

\bibitem{NIPS2014_5542}  Mnih, V., Heess, N., Graves, A.: Recurrent models of visual attention. NIPS 2014.

\bibitem{GlobalPooling} Lin, M., Chen, Q. and Yan, S.: Network in network. ICLR, 2014.

\bibitem{mincostflow} Papadimitriou, C.H., Steiglitz, K.: Combinatorial Optimization: Algorithms and Complexity. Dover Publications, Inc., 1988.

\bibitem{ksp} Berclaz, J., Fleuret, F., Turetken, E., Fua, P.: Multiple Object Tracking Using K-Shortest Paths Optimization, IEEE PAMI 
Vol. 33, No. 9,  p.1806-1819, September 2011.



\end{thebibliography}
\end{document}